\documentclass[conference]{IEEEtran}
\IEEEoverridecommandlockouts
\usepackage{cite}
\usepackage{amsmath,amssymb,amsfonts}
\usepackage{algorithmic}
\usepackage{graphicx}
\usepackage{textcomp}
\usepackage{xcolor}
\usepackage{balance} % for balancing columns on the final page
\usepackage{comment}
\usepackage{colortbl}
\usepackage{subfig}
\usepackage[fleqn]{nccmath}
\usepackage{siunitx}
\usepackage{url}
\usepackage{balance}
\usepackage{fancyhdr}
\usepackage{ragged2e}

\def\BibTeX{{\rm B\kern-.05em{\sc i\kern-.025em b}\kern-.08em
    T\kern-.1667em\lower.7ex\hbox{E}\kern-.125emX}}
\begin{document}

\fancypagestyle{firstpage}
{
    \fancyhf{}

    \fancyhead[L]{\small \justifying \noindent Preprint accepted at the IEEE/RSJ International Conference on Intelligent Robots and Systems (IROS), Kyoto, Japan, 2022. 
    }

}

\title{Evaluating Human-like Explanations for Robot Actions in Reinforcement Learning Scenarios\\
%\thanks{Francisco Cruz gratefully thanks financial support from the Faculty of Science, Engineering \& Built Environment of Deakin University through the Peer-Review Early Career Researchers Support Scheme (PRESS).
%Charlotte Young was supported by an Australian Government Research Training Program (RTP) Stipend and RTP Fee-Offset Scholarship through Federation University Australia.}
}

\author{\IEEEauthorblockN{
Francisco Cruz\,$^{1,4}$,
Charlotte Young\,$^{3}$,
Richard Dazeley\,$^{2}$, and
Peter Vamplew\,$^{3}$
}
\IEEEauthorblockA{\,
$^{1}$School of Computer Science and Engineering, University of New South Wales, Sydney, Australia\\
$^{2}$School of Information Technology, Deakin University, Geelong, Australia\\
$^{3}$School of Engineering, IT and Physical Sciences, Federation University, Ballarat, Australia\\
$^{4}$Escuela de Ingenier\'ia, Universidad Central de Chile, Santiago, Chile\\
Emails: f.cruz@unsw.edu.au, richard.dazeley@deakin.edu.au, \{cm.young, p.vamplew\}@federation.edu.au
}
}

\maketitle

\begin{abstract}
Explainable artificial intelligence is a research field that tries to provide more transparency for autonomous intelligent systems.
Explainability has been used, particularly in reinforcement learning and robotic scenarios, to better understand the robot decision-making process.
Previous work, however, has been widely focused on providing technical explanations that can be better understood by AI practitioners than non-expert end-users.
In this work, we make use of human-like explanations built from the probability of success to complete the goal that an autonomous robot shows after performing an action.
These explanations are intended to be understood by people who have no or very little experience with artificial intelligence methods. 
This paper presents a user trial to study whether these explanations that focus on the probability an action has of succeeding in its goal constitute a suitable explanation for non-expert end-users.
The results obtained show that non-expert participants rate robot explanations that focus on the probability of success higher and with less variance than technical explanations generated from Q-values, and also favor counterfactual explanations over standalone explanations.
\end{abstract}


\thispagestyle{firstpage}

\section{Introduction}

Artificial Intelligence (AI) models are usually designed by phenomenological rules, empirical observation, or as a combination of both.
These models are also known as white, black, and gray-box models, respectively~\cite{cruz2007indirect, cruz2010indirect}. 
Explainable Artificial Intelligence (XAI) is a field of study aimed at reliably and efficiently capturing the AI decision-making process, and interpreting and reporting it to a human audience~\cite{dazeley2021levels}.

In Reinforcement Learning (RL) and robotic scenarios, eXplainable RL (XRL) has been mostly focused on explanations built from the perceived robot's state, not considering the underlying goals to complete a task~\cite{anjomshoae2019explainable}.
Moreover, agents have often used technical jargon to explain its behavior, e.g., `I chose this action because it maximizes future collected reward' or `I chose that action because it is the next one following the optimal policy'~\cite{degraff2017people}. 
%However, these kinds of explanations are not adequate for non-expert end-users.

\begin{comment}
Unfortunately, it is still unclear how to judge if an explanation is adequate for non-expert end-users or how to correctly evaluate a good explanation~\cite{doshi2017towards}.
For instance, previous works have used technical language to explain agent's behavior~\cite{pocius2019strategic, lengerich2017towards, juozapaitis2019explainable}. 
These explanations are in principle directed to machine learning practitioners, but non-experts will have difficulty in understanding these explanations.
Using goal-driven explanations, the probability of success has been previously used trying to explain the agent's behavior to non-expert end-users~\cite{cruz2019memory}.
However, the use of the probability has not been quantified in terms of explainability understanding benefits for end-users.
\end{comment}

Unfortunately, it is still unclear how to judge if an explanation is adequate for non-expert end-users or how to correctly evaluate a good explanation~\cite{doshi2017towards}. 
For instance, previous works have used technical language to explain agent's behavior to end-users, e.g. salience maps~\cite{greydanus2018visualizing}, reward decomposition~\cite{juozapaitis2019explainable}, and graphs~\cite{madumal2020explainable}.
These explanations might be in principle directed to machine learning practitioners, but non-experts might find difficulties understanding these explanations. 
Using goal-driven explanations, the probability of success has been previously used in trying to explain the agent's behavior to non-expert end-users~\cite{cruz2019memory}. 
However, the use of this probability has not been quantified in terms of understanding explainability benefits for end-users.

In this work, we study how non-expert end-users rate human-like explanations, in the form of `action $a_i$ gives an $x_i$\% chance of success the task compared to $x_j$\% for action $a_j$'.
We carried out a human trial to investigate if human-like explanations using the probability of success may better explain robot’s actions to non-expert end-users than technical explanations.
To this end, we have created four different scenarios in which participants observe an initial and final situation after a robot performs an action.
%We present empirical results on how good the perceived explainability is after performing a particular action in the robotic scenarios.

When a scenario was presented to a participant, they were also randomly provided with either a technical explanation (created from the Q-values) or a human-like explanation (created from the computed probability of success). 
Both of these explanations are computed by an explainable RL approach.
A total of 228 users have evaluated the proposed scenarios and ranked standalone and counterfactual explanations. % through the Amazon Mechanical Turk platform.
Results obtained show how understandable an explanation created from the probability of success is, demonstrating it as a good basis for human-like explainability.

\section{Explainable reinforcement learning} \label{sec:xrl_approaches}

Explainable Reinforcement Learning (XRL) aims to extract explainable concepts from the learning agent using environmental perception, intrinsic and extrinsic motivations, and beliefs~\cite{dazeley2021explainable}.
Explanations may be directed to machine learning practitioners, in which case explanations are often given using technical language to explain the agent's behavior.
These explanations are mainly based on perceived features from the state (i.e., state-based explanations) and they do not address goal-driven action explainability~\cite{puiutta2020explainable}.

For example, salience maps have been previously used to provide visual explanations of a learning agent using deep RL~\cite{pocius2019strategic}, or distal explanations by means of decision trees and causal models~\cite{madumal2020distal}.
Furthermore, a hybrid approach for interpretable policies mixing RL with genetic programming has been used to explain policies using equations to this aim~\cite{hein2018interpretable}, or programmatically interpretable RL has been introduced to explain agent's behavior using only symbolic inputs ~\cite{verma2018programmatically}. 
Reward decomposition has been also utilized in order to explain to RL practitioners the agent's action selection~\cite{juozapaitis2019explainable}.

For sequential decision-making problems, usually there is no clear guidance on what makes a good explanation and, for instance, for an agent using Q-learning an explanation might take the form of ``the action had the highest Q-value given this state'' \cite{ehsan2018rationalization}.
Algorithm-based explanations are particularly tailored for specific algorithms and this method has become quite common in explaining decisions generated by deep reinforcement learning~\cite{chakraborti2020emerging}.
For example, authors in~\cite{tabrez2019explanation} use a preestablished sentence in the form ``If you perform \{describe\_action($\pi_h$)\}, you will fail the task in state \{describe\_state($s_{h,terminal}$)\} because of \{describe\_reward(diff($R_h$,$R_{\pi}$))\}”. 
Thus, this explanation still depended on a technical value returned by the describe\_reward function.

In case of explanations directed to non-experts, the probability of successfully completing the intended task after performing an action has been previously used~\cite{barros2020moody, cruz2021explainable}. 
In this regard, three methods have been proposed in order to compute the probability of success of a learning robot using RL to perform a task -- namely, memory-based, learning-based, and introspection-based methods. 
In this work, we make use of the computed probability of success with these methods in order to create an explanation of the robot decision-making process and exhibit it to a non-expert end-user to evaluate it.

%\begin{comment}

The memory-based method uses an episodic memory to save all the robot state-action transitions using a list.
Once the robot reaches the goal position, the probability of success is computed according to Eq. \eqref{eq:memory-based} as follows:

\begin{equation}
    P_s \leftarrow T_s/T_t,
\label{eq:memory-based}
\end{equation}
where 
$P_s$ is the probability of success, $T_t$ is the total number of transitions, and $T_s$ is the number of transitions in a success sequence.

The learning-based method learns in parallel the probability of success by either using an additional table or a function approximator.
Therefore, this method iteratively updates the so-called $\mathbb{P}$-values to represent the probability of success associated with a state-action pair.
The probability is computed according to Eq. \eqref{eq:learning-based} as follows:

\begin{equation}
\mathbb{P}(s_t,a_t) \leftarrow \mathbb{P}(s_t,a_t) + \alpha [ \varphi_{t+1} + \mathbb{P}(s_{t+1},a_{t+1}) - \mathbb{P}(s_t,a_t) ],
\label{eq:learning-based}
\end{equation}
where $a$ and $s$ are the action and the state, respectively. 
$\mathbb{P}$ is the probability of success associated with a state-action pair, i.e. the $\mathbb{P}$-value. 
$\alpha$ is the learning rate and $\varphi$ is a success flag used when the task is finished.

The introspection-based method estimates the probability of success directly from the Q-values and, therefore, no additional memory is needed.
The estimated probability of success $\hat{P}_s$ is computed according to Eq. \eqref{eq:introspection-based} as follows:

\begin{ceqn}
\begin{align}
\hat{P}_s \approx \left[ \frac{1}{2} \cdot log_{10} \frac{Q(s,a)}{R^T} + 1 \right]^{\hat{P}_s \le 1}_{\hat{P}_s \ge 0},
\label{eq:introspection-based}
\end{align}
\end{ceqn}
where the $Q(s,a)$ is the state-action pair, and $R^T$ is the total reward obtained when the agent finishes the task.
Additionally, to properly shape the probability a logarithmic transformation is performed and the obtained value is bounded as $\hat{P}_s \in [0,1]$.

%\end{comment}

\section{Experimental methodology}
In this section, we present the details of the research method used to collect data and for posterior analysis.
We introduce the platform utilized for recruiting volunteers, and describe %the demographic data collected, and 
how the scenarios are presented and ranked during the explainability experiments\footnote{The present study has received ethics approval by Deakin University (reference number: SEBE-2021-29).}.

\subsection{Survey platform}\label{sec:survey_platform}
Amazon Mechanical Turk (MTurk) %\footnote{https://www.mturk.com/} 
is a crowdsourcing marketplace in which a requester may hire remotely located workers to solve a task.
The distributed workforce address the job, usually referred to as a Human Intelligence Task (HIT), and receives compensation for doing the job.
HITs may be, for example, answering a questionnaire, labeling data, or writing descriptions~\cite{litman2020conducting}. 
Although there has been some discussion about the benefits of crowdsourcing platforms regarding the quality of data provided by subjects~\cite{paolacci2010running} or regarding the quality of instructions given to the crowd~\cite{wu2017confusing}, MTurk has shown to be an important complement to existing methods for recruiting participants in research experiments~\cite{strickland2019use}, especially nowadays considering the reduction of many in-person activities after COVID-19 interruption~\cite{moss2020demographic}.

In this work, we used the MTurk platform in order to recruit participants for our study.
Involvement in the study was voluntary and each worker received compensation of USD \$2.50 through MTurk upon completion.
We allocated 20 minutes for the HIT leading to a hourly payment of USD \$7.50. 
Although the estimated time to answer the survey was 12 minutes, we allowed extra time as previously suggested by Navarro~\cite{navarro2015ethical}.
On the MTurk platform we requested workers with master qualifications.
A master worker has previously demonstrated good performance in a wide range of tasks, a large number of times.
Particularly for our survey, we have recruited workers with HIT approval rates greater than or equal to 95\%.
The request for master qualifications cost an additional 5\% per worker. This fee was paid to the MTurk platform itself, not the workers.

A possible limitation of using MTurk in our study is that usefulness might not be the main axis of concern as participants did not perform the tasks. 
Instead, we showed them images and videos of initial and final situations along with an explanation for their evaluation.
However, reinforcement learning tasks are well-known for being very time-demanding and as soon as we increase considerably the time needed for the survey, the number of participants would have decreased dramatically (as we would need to pay a higher fee to each of them) as well as the quality of the answers being affected especially for the final scenarios as a matter of patience.

\subsection{Evaluating explanations in robot scenarios}

As this study focuses on the evaluation, by non-expert end-users, of explanations that use the probability of success, we initially collected information to better understand and identify demographic characteristics of the participants.
In this regard, the main question used for analysis was \textit{How would you rate your level of expertise in machine learning?}.
The possible answer was from 0 (beginner) to 10 (expert).

In order to test that an explanation, that we created from the probability of success, is a suitable basis to explain robot behavior, we implemented four robotic scenarios. 
Each scenario changes the situation the robot is in after performing an action.
Each scenario explains the robot behavior using either a technical explanation, created from the Q-values, or an human-like explanation, created from the computed probability of success.
The explanation used is selected randomly each time.

Moreover, to explain the robot's actions, we use both standalone and counterfactual explanations.
In case the robot is directly asked about its last performed action, we use standalone explanations, i.e., the robot explains its behavior by directly answering the question in terms of the probability of success or Q-values.
When counterfactual questions are asked, e.g., 'why you did not perform a specific action?', the robot explains what it did as contrasting two possible actions (the one asked and the one performed).

All scenarios exhibit the same presentation format as follows:

\begin{itemize}
    \item Brief introduction of the scenario, the robot goal, and possible actions.  %\todo{Should I include the reward function for each?}.
    \item Initial situation before performing a particular action into the environment.
    \item Final situation after performing the previous action.
    \item Question asked to the robot about its previous behavior (direct or counterfactual).
    \item Robot's answer to the previous question using either technical or probability-based language.
    \item Evaluation of the provided explanation for the robot's action. For this, we asked participants: How would you rate this explanation to the robot's action? (between 0 as not at all useful and 10 as extremely useful).
\end{itemize}

The robot's answer provides an explanation for the robot's action, however, each participant receives only one randomly generated explanation using either a Q-value or an estimated probability of success.
In both cases, the robot explanation is generated with the knowledge the robot has collected.
In the case of using the probability of success, the robot uses actual experimental data computed with the memory-based and introspection-based methods proposed by~\cite{cruz2021explainable}.
For the evaluation, participants are allowed to rate the provided explanations in terms of their usefulness.
The rating scale is defined from 0, meaning not useful at all, to 10, meaning extremely useful.

In our study, we asked participants to evaluate eight different explanations.
The robot actions to be explained were divided into four scenarios with two explanations for each, i.e., two robot actions leading to different explanations respectively.
The scenarios were shown always in the same order from the simplest grid-world environment to the real robot environment, i.e., the island scenario, the robot navigation scenario, the robot arm scenario, and the real-world Nao robot scenario.
However, within the same scenario, each of the two possible robot actions was shown randomly to participants in order to avoid bias by previously experiencing the same scenario.
Moreover, after observing the final state yielded from the robot action, each participant received randomly either a technical or a probability-based explanation for each situation as well.
In the next section, we explain the details for each of the scenarios.

\begin{comment}
The scenarios are divided into four groups: the island scenario, the robot navigation scenario, the robot arm scenario, and the real-world Nao robot scenario.
Each of these four groups contains two different situations to explain, i.e., two initial situations leading to two different final situations after performing an action.
Each of these two situations within a group is shown randomly to participants in order to avoid bias by previous experiencing the same scenario.
In the next section, we explain the details for each of the scenarios. 
\end{comment}

\section{Experimental robotic scenarios}

In this section, we introduce the robotic scenarios used to generate the explanations. 
%The explanations produced for each scenario may subsequently be ranked, by the participants, by their usefulness.
As previously mentioned, we created four different scenarios with two robot actions each.
For the sake of space, in this work, only one robot action for each scenario is shown depicting the initial and the final situation. 
However, all situations are described and explained below\footnote{The full survey, including all situations and figures, may be accessed at \url{https://researchsurveys.deakin.edu.au/jfe/form/SV_5dPVDr9171GCUQe}.}. 
Regardless of the underlying RL representation of each scenario, i.e., state and action space,
%and 
discretized or continuous domain, 
or the action selection method~\cite{cruz2018action},
in this work, we are not focused on how the agent learns the solution but rather to provide and evaluate explanations given to non-expert end-users of what motivates the agent's specific actions from different states in different domains.

\subsection{Island scenario}
\begin{figure}[htbp]
\centering
\subfloat[Initial state.]{\includegraphics[width=0.5\linewidth]{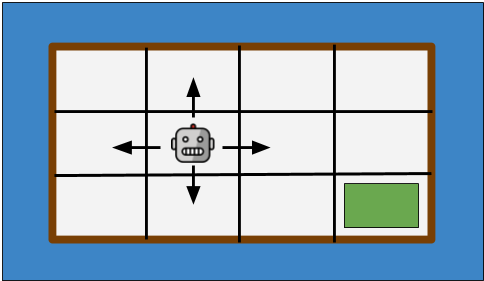} 
\label{fig:island_scenario_1}
}
%\\
\subfloat[State after `go east' action.]{\includegraphics[width=0.5\linewidth]{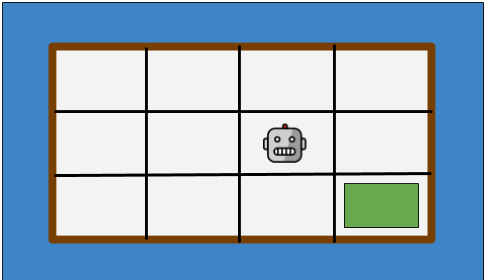} 
\label{fig:island_scenario_2}
}
\caption{Island scenario. In this situation, the robot performs the action  `go east' into the grid world island scenario.}
\label{fig:island_scenario}
\end{figure}

Imagine a square-shaped island. 
In this scenario, a robot needs to navigate toward the goal from a random initial position.
To do this, four movements are possible: go north, go south, go east, and go west. 
The island scenario is in fact a mock-up from a simple real grid-world RL scenario.
The problem has been previously addressed by Cruz et al.~\cite{cruz2019memory} using memory-based explainable reinforcement learning (MXRL).
Figure~\ref{fig:island_scenario} shows the island scenario in which the goal position is depicted in green.

As mentioned, two situations are presented in this scenario.
The first one is shown in Figure~\ref{fig:island_scenario}.
In this case, the robot is initially placed at position $(1,1)$ (shown in Figure~\ref{fig:island_scenario_1}) and, after performing the action `go east', the robot finishes at position $(2,1)$ (shown in Figure~\ref{fig:island_scenario_2}).
After performing `go east', the robot is asked: `why did you move to the east?'. 
The robot may answer using technical or probability-based language as: %'I moved to the east because it has a Q-value of -0.411' or `I moved to the east because it has a 65.6\% probability of reaching the green position' using technical or probability-based language respectively.
\begin{itemize}
    \item `I moved to the east because it has a Q-value of -0.411', or 
    \item `I moved to the east because it has a 65.6\% probability of reaching the green position'
\end{itemize}

In the second situation (not shown graphically in this paper), the robot is initially placed at position $(3,0)$ and, after performing the action `go south', the robot finalizes at position $(3,1)$.
After performing `go south', the robot is asked: `why you did not move to the east?'.
The robot may explain using technical or probability-based language as: 

\begin{itemize}
    \item `I did not move to the east because it has a Q-value of -0.998, while moving south has a Q-value of 0.181', or
    \item `I did not move to the east because it has 0\% probability of reaching the green position, instead moving south has 73.6\% probability'
\end{itemize}

%'I did not move to the east because it has a Q-value of -0.998, while moving south has a Q-value of 0.181' using technical language.
%Alternatively, the robot may explain: `I did not move to the east because it has 0\% probability of reaching the green position, instead moving south has 73.6\% probability'.
%Once again, the participant ranks the explanation for the robot's action using a scale from 0 to 10.

\subsection{Robot navigation scenario} \label{sec:navigation_scenario}
\begin{figure}[htbp]
\centering
\subfloat[Initial state.]{\includegraphics[width=0.5\linewidth]{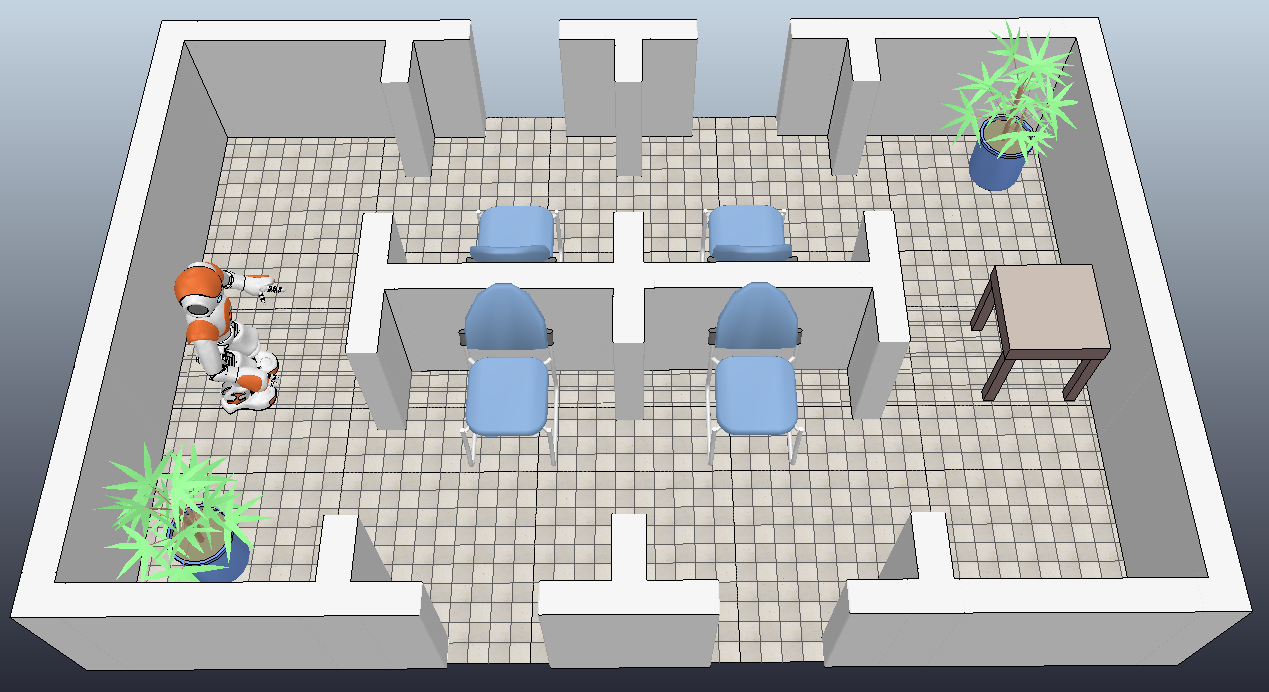} 
\label{fig:navigation_scenario_1}} 
%\hspace{0.5cm}
%\\
\subfloat[State after `move right' action.]{\includegraphics[width=0.5\linewidth]{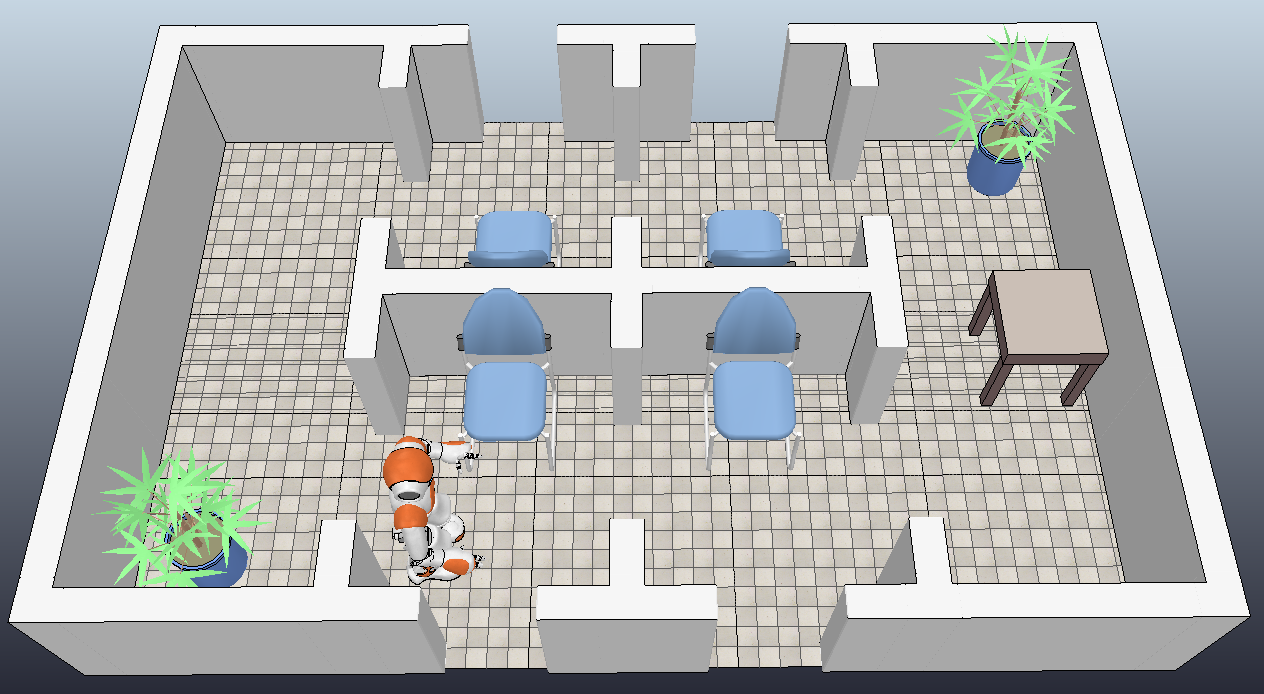} 
\label{fig:navigation_scenario_2}
}
\caption{Navigation scenario. In this situation, the robot performs the action `move to the right' from the initial position.}
\label{fig:navigation_scenario}
\end{figure}

Consider a robot navigating through different rooms trying to reach a table, as shown in Figure~\ref{fig:navigation_scenario}.
In the robot navigation scenario, a robot is placed in a fixed initial position and needs to reach the table to the right (see Figure~\ref{fig:navigation_scenario}). 
In this scenario, four actions are possible from the robot's perspective: move to its left, move to its right, move straight, or stay in the same room. 
This scenario has been developed with the CoppeliaSim %\footnote{https://www.coppeliarobotics.com/} 
robot simulator given the good performance previously shown~\cite{ayala2020comparison}.
A similar scenario has been previously tested with learning-based and introspection-based explainable reinforcement learning methods~\cite{cruz2021explainable}.

The first situation tested with this scenario is depicted in Figure~\ref{fig:navigation_scenario}.
In this situation, the robot is just starting the task, therefore, it is placed at the initial position to the left of the scenario (shown in Figure~\ref{fig:navigation_scenario_1}).
After performing the action `move right', the robot finishes in an intermediate room as shown in Figure~\ref{fig:navigation_scenario_2}.
Once the action is completed, the robot is asked: `why did you move to the right?'.
The robot explains this behavior by answering either using technical or probability-based language as:

\begin{itemize}
    \item `I moved to the right because it has a Q-value of 0.621', or
    \item `I moved to the right because it has an 89.7\% probability of reaching the table'
\end{itemize}

In the second situation in this scenario (not shown graphically here), the robot is currently placed in an internal room, just one action away from either finishing the task or moving out from the scenario, with the latter leading to a negative reward and restarting of the learning episode.    
After performing the action `move straight', the robot finishes in the goal room reaching the table.
The robot is then asked: `why you did not move to the right?'.
The robot may explain its behavior referring to the technical or probability-based explanation as: 

\begin{itemize}
    \item `I did not move to the right because it has a Q-value of -1, while moving straight has a Q-value of 1', or
    \item `I did not move to the right because it has a 0\% probability of reaching the table, while moving straight has a 100\% probability'
\end{itemize}

\subsection{Robot arm scenario}
\begin{figure}[htbp]
\centering
\subfloat[Initial state.]{\includegraphics[width=0.5\linewidth]{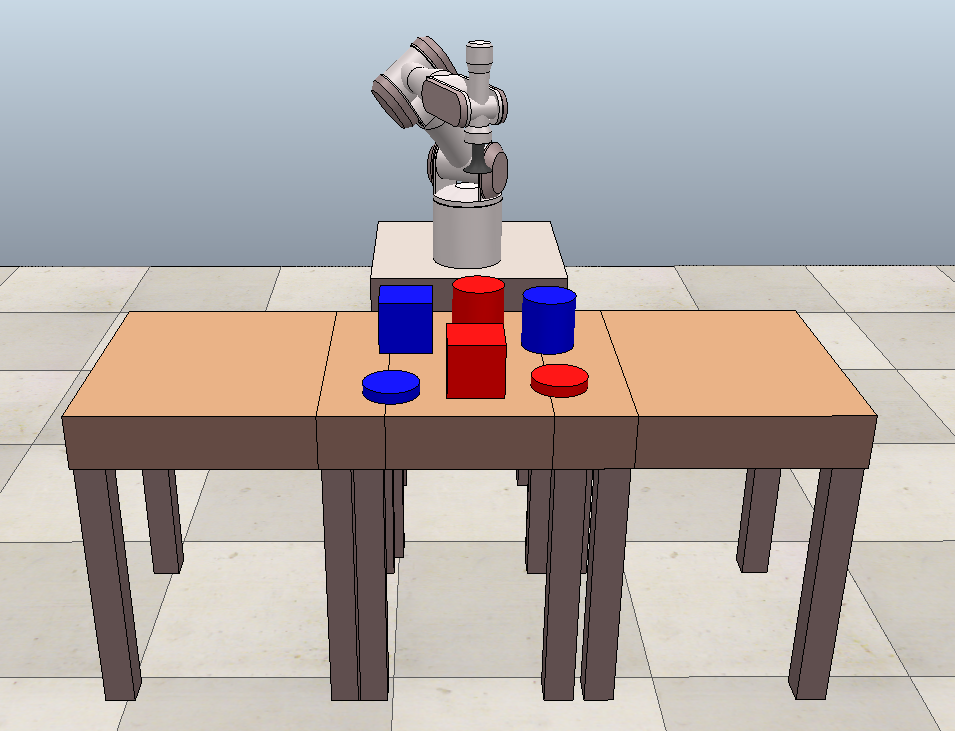} 
\label{fig:robot_arm_scenario_1}
} 
%\hspace{1cm}
%\\
\subfloat[State after `grab an object' action.]{\includegraphics[width=0.5\linewidth]{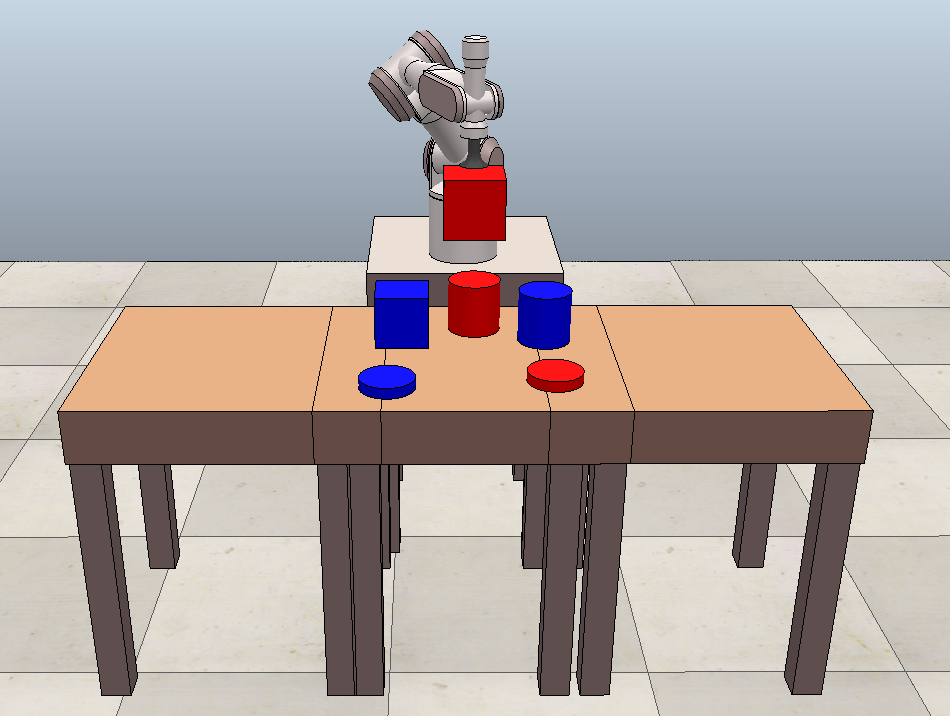} 
\label{fig:robot_arm_scenario_2}
}
\caption{Robot arm scenario. In this situation, the robot performs the action `grab an object'.}
\label{fig:robot_arm_scenario}
\end{figure}

Let us now consider a robot arm that has to sort different objects according to their color, so that red objects are placed on a table to the right of the arm and blue ones to the left, as depicted in Figure~\ref{fig:robot_arm_scenario}.
The robot arm scenario comprises six different objects that need to be sorted as described above.
Initially, all the objects are placed on a table in the middle of the scenario.
At each moment, the robot can perform four different actions: grab an object, move to the right, move to the left, and drop an object.
By performing these actions, the robot may pick up and move the objects in order to sort them by color.
This scenario has been also developed using the CoppeliaSim robot simulator as in the previous scenario.
This continuous scenario was originally proposed by Moreira et al.~\cite{moreira2020deep} and subsequently used for a robust interactive actor-critic approach~\cite{millan2021robust} and for explainability in robotic systems~\cite{cruz2021explainable}.

In this scenario, the first situation places the robot arm at the initial position, i.e., at the center of the table upon the unsorted objects, as shown in Figure~\ref{fig:robot_arm_scenario_1}.
In this case, the action `move to the right' is performed and the robot arm finishes above the table to the right (not shown graphically here). 
After performing the action, the robot is asked: `why did you move to your right?'.  
The robot may explain its behavior using either technical or probability-based jargon respectively as:

\begin{itemize}
    \item `I moved to the right because this action has a Q-value of 0.237', or
    \item `I moved to the right because by doing so I have 55.1\% probability of completing the task successfully'
\end{itemize}

The second situation in this scenario starts from the same initial position as the previous case, i.e. at the center of the table upon the unsorted objects, as shown in Figure~\ref{fig:robot_arm_scenario_1}.
After performing the action `grab an object', the robot grabs randomly one of the objects and finishes with a red cube picked up, as shown in Figure~\ref{fig:robot_arm_scenario_2}.
After the action execution, the robot is asked the following question: `why did you not move to your left?'.
The robot explains its behavior using technical or probability-based language as:

\begin{itemize}
    \item `I did not move to the left because it has a Q-value of 0.161, while grabbing an object has a Q-value of 0.465', or
    \item `I did not move to the left because it has a 38.4\% probability of finishing the task successfully, while grabbing an object has a 60.8\% probability'
\end{itemize}

\subsection{Real-world Nao robot scenario}

\begin{figure}[htbp]
\centering
\subfloat[Initial state.]{\includegraphics[width=0.5\linewidth]{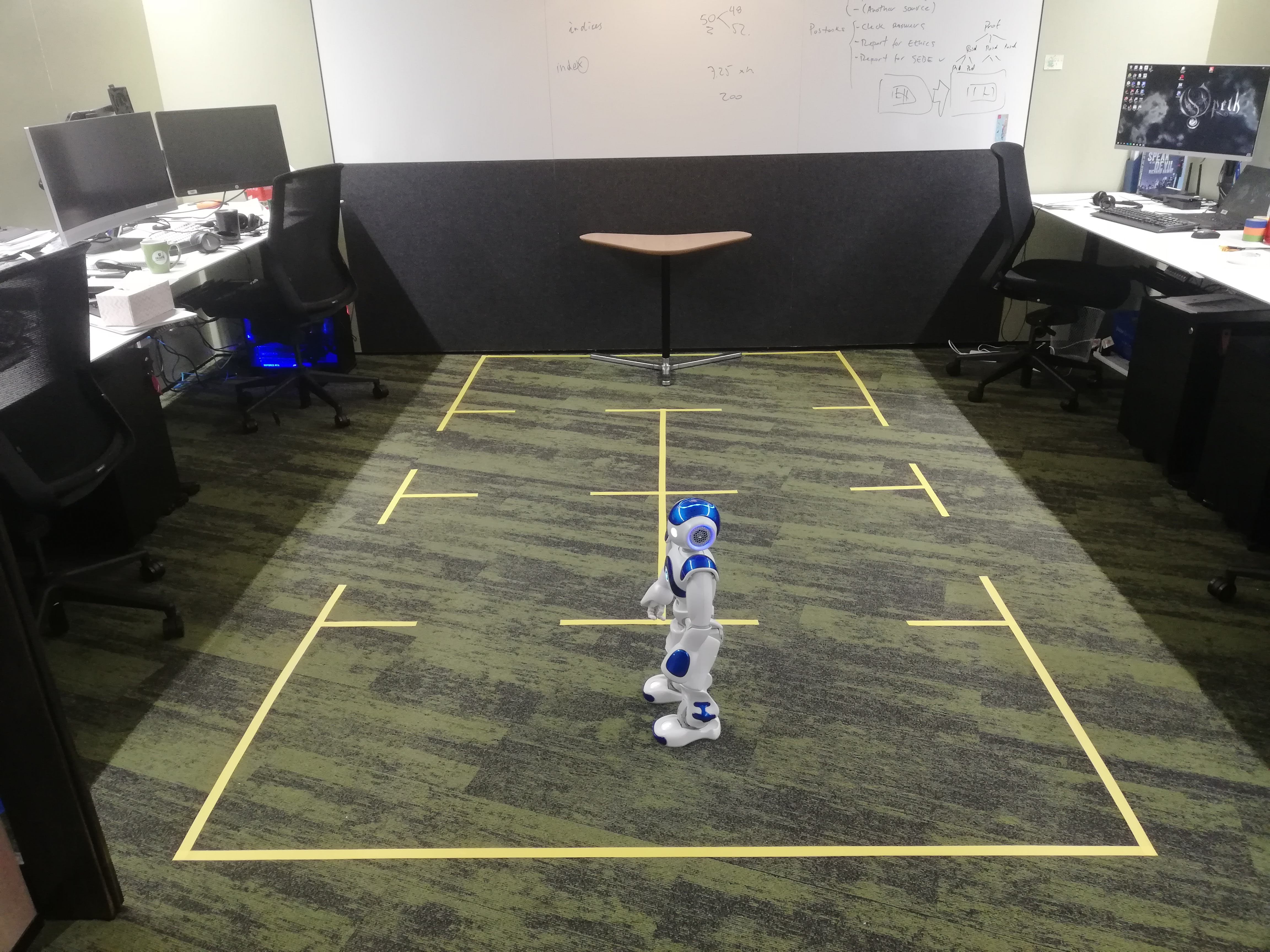} 
\label{fig:real_nao_scenario_1}
} 
%\hspace{1cm}
%\\
\subfloat[State after `move to the left' action.]{\includegraphics[width=0.5\linewidth]{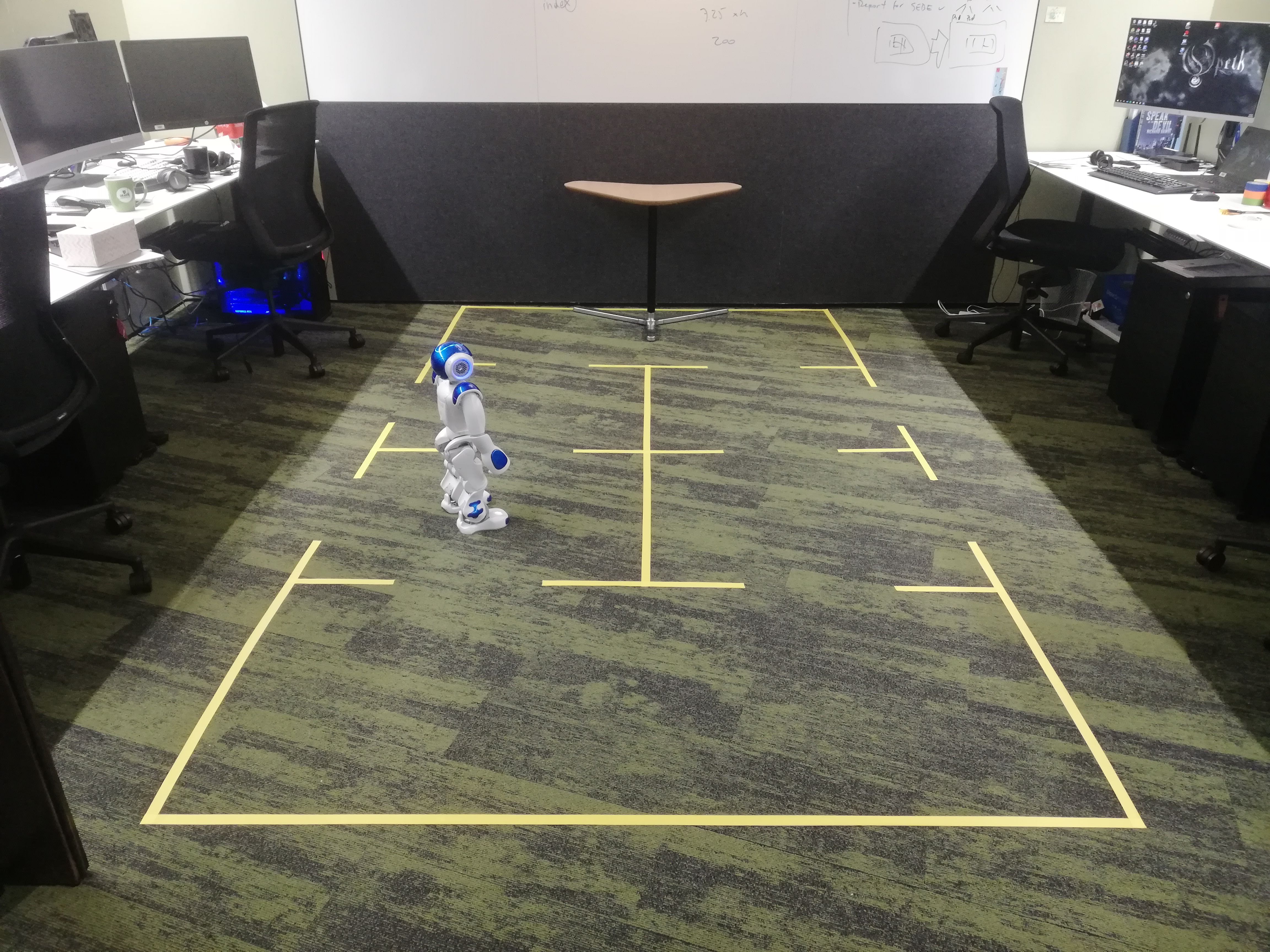} 
\label{fig:real_nao_scenario_2}
}
\caption{Real-world Nao robot scenario. In this situation, a real Nao robot performs the action `move to the left' from the initial position.}
\label{fig:real_nao_scenario}
\end{figure}

Consider a robot navigating through different sections on the floor trying to reach the little table at the back, as shown in Figure~\ref{fig:real_nao_scenario}. 
This scenario is a real-world implementation of the simulated robot navigation task described in Section~\ref{sec:navigation_scenario}.
Therefore, the robot's goal and the state-action representation are exactly the same. % as before, i.e. to reach the table as navigating through the internal rooms from a fixed initial position.
%In this real-world scenario, the robot also chooses between four possible actions from its perspective: move to the left, move to the right, move straight, and stay in the same section. 
In this case, we have recorded videos using a real Nao robot to show the situation to participants. %\footnote{The video recordings are available at \url{https://youtu.be/xu47_1SNyEA} and \url{https://youtu.be/pxuMxx0jkek}.}. %\url{https://youtu.be/xu47_1SNyEA}.}).
%The recordings are publicly available for further evaluation.

The first situation in this scenario is the robot starting from an intermediate room close to the table (this situation is not shown here with images). 
After performing the action `move straight', the robot finishes in the goal position next to the table.
Then the robot is asked: `why did you move straight?', in order to explain its previous behavior.
The robot explains its behavior by using either technical or probability-based language as follows: 

\begin{itemize}
    \item `I moved straight because it has a Q-value of 1', or
    \item `I moved straight because it has a 100\% probability of reaching the table'
\end{itemize}

The second situation in this scenario shows the robot starting from the initial position and performing the action `move to the left', as shown in Figure~\ref{fig:real_nao_scenario_1}. 
%\footnote{Recording for this situation available at [hidden url].}.%\url{https://youtu.be/pxuMxx0jkek}.}.
After finishing the action execution, the robot moves to the next internal room to the left, as shown in Figure~\ref{fig:real_nao_scenario_2}.
The robot is then asked to explain its behavior as: `why you did not move to the right?'.
The two possible explanations given by the robot using technical and probability-based language respectively are: 

\begin{itemize}
    \item `I did not move to the right because it has a Q-value of 0.621, while moving to the left has a Q-value of 0.744', or
    \item `I did not move to the right because it has a 89.7\% probability of reaching the table, while moving to the left has a 95.8\% probability'
\end{itemize}

\section{Results}

Originally, we collected answers from 228 participants. As previously mentioned in Section \ref{sec:survey_platform}, the estimated time needed to complete the survey was twelve minutes. 
We observed that about 80\% of participants spent on average $12 \pm 5$ minutes to finalize all the survey questions and roughly 20\% less than five minutes. 
We considered that these participants might have not taken enough care with their answers and we decided not to include them. 
Thus, 183 participants' answers were included in the analysis.

\begin{comment}
%In this section, we present the results obtained from the survey.
Originally, we collected answers from 228 participants, however, 45 participants completed the survey in less than five minutes with inconsistent answers.
We considered that these participants did not take enough care with their answers, and so we did not include their answers in our analysis.
%We consider these participants as not taking enough attention to the questions and their answers were not included in the analysis.
Thus, 183 participants' answers were included in the analysis.
\end{comment}

\begin{figure}[htbp]
\centering
\includegraphics[width=0.7\linewidth]{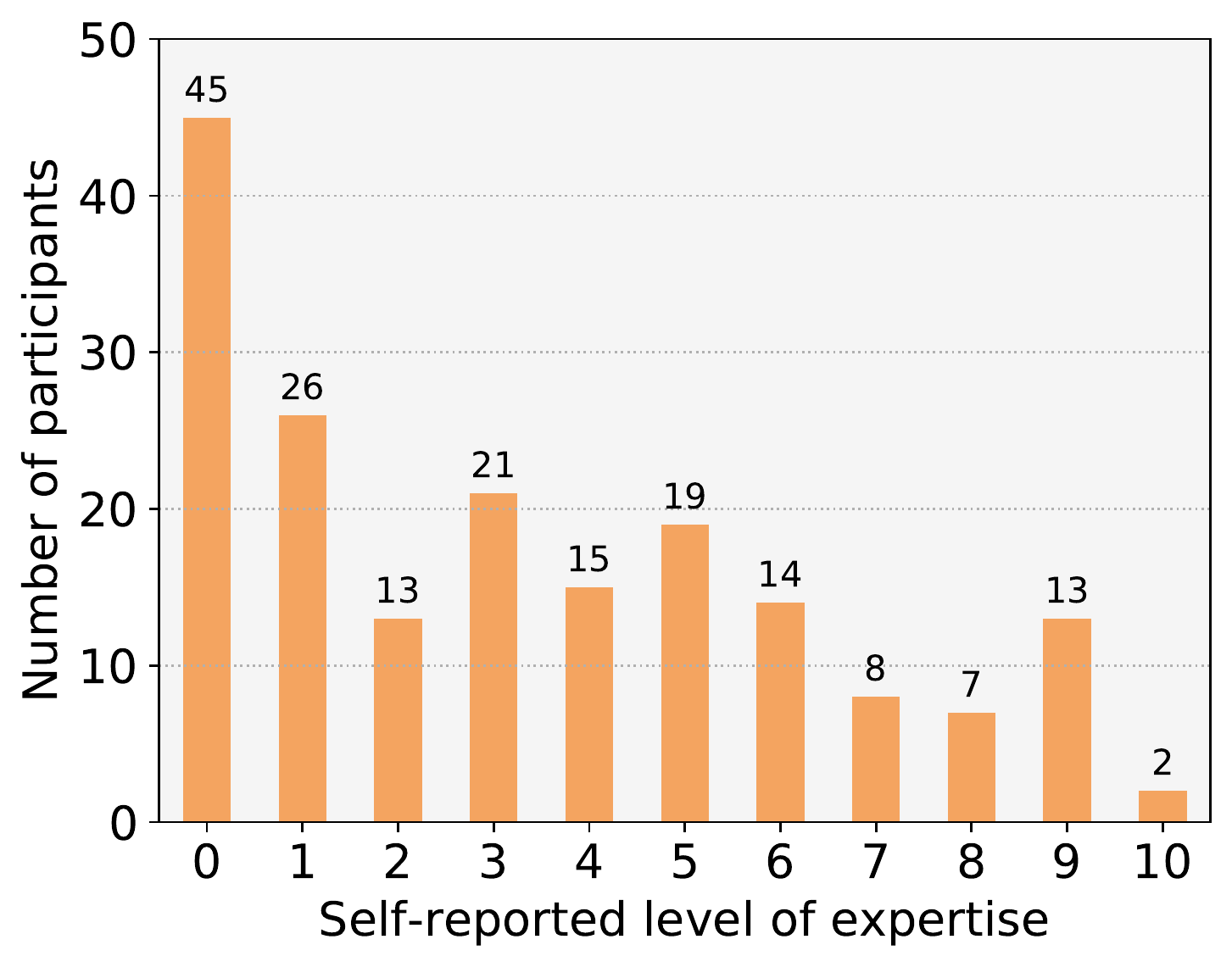} 
\caption{Self-reported level of expertise in machine learning. Most participants reported no previous experience in machine learning. The group of participants reporting 0, 1, and 2 as levels of expertise are referred to as non-expert end-users.}
\label{fig:expertise}
\end{figure}

\begin{figure*}[htbp]
\centering
\subfloat[Island scenario -- Go east.]{\includegraphics[width=0.25\linewidth]{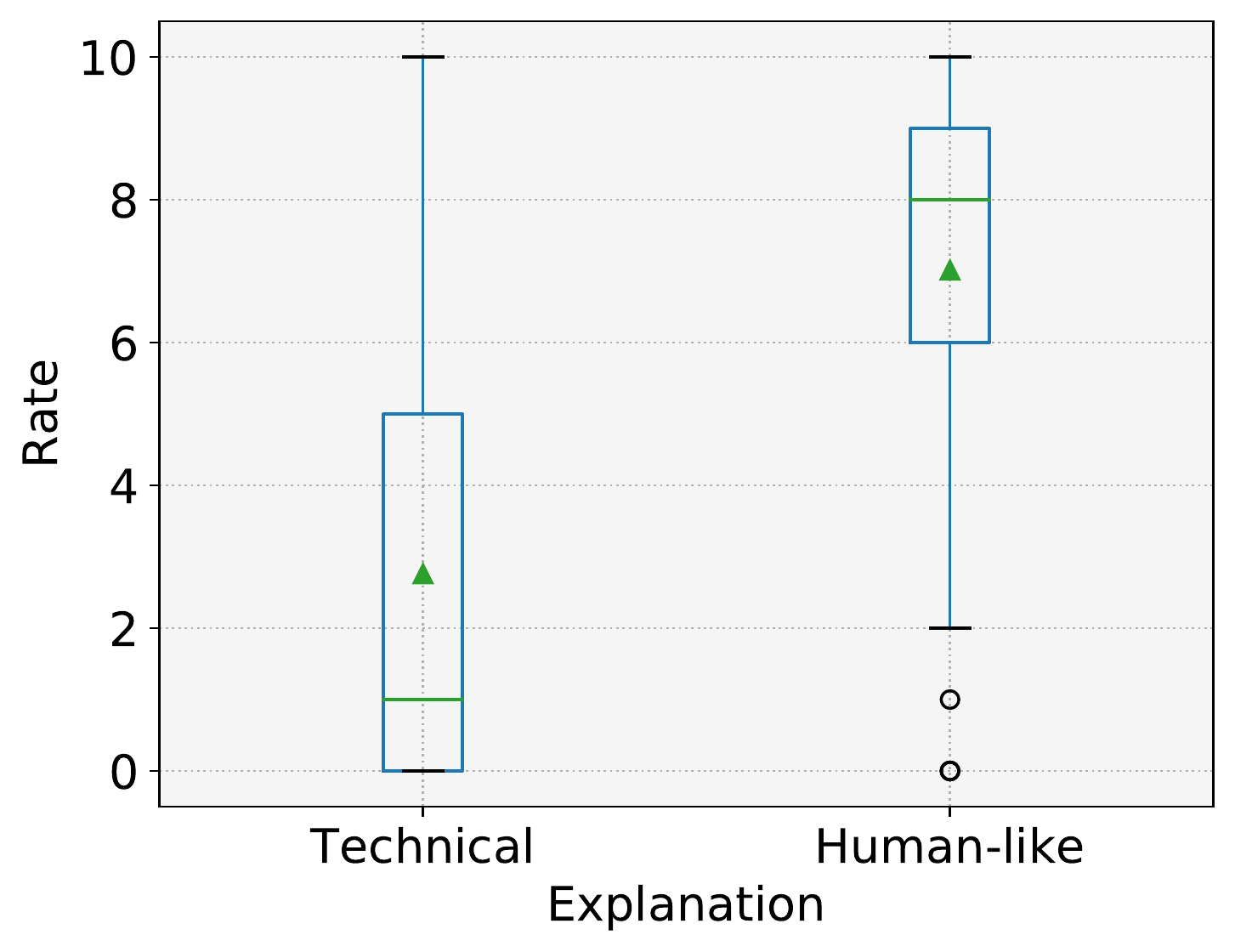} 
\label{fig:scenario_1_results}
} 
\subfloat[Island scenario -- Go south.]{\includegraphics[width=0.25\linewidth]{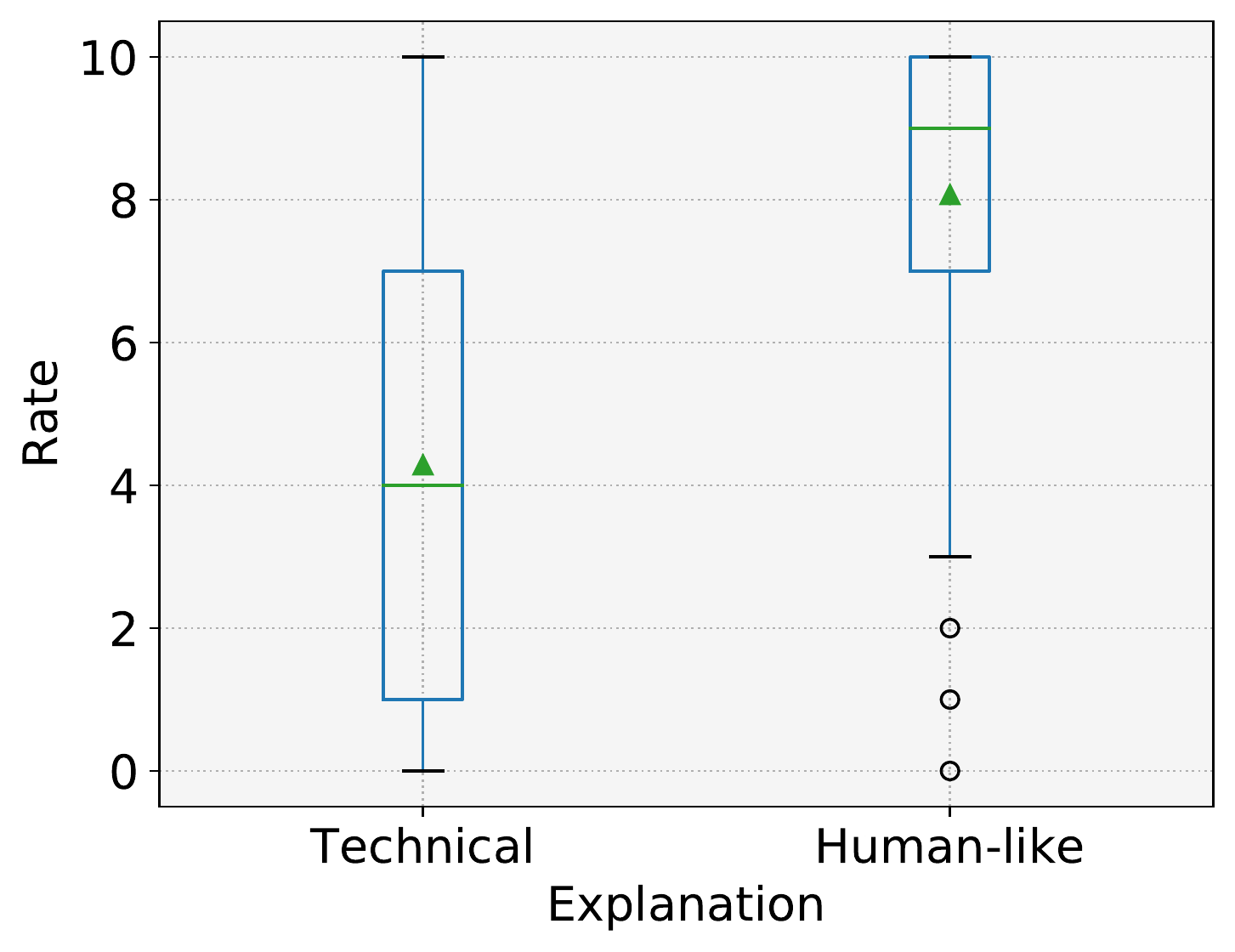} 
\label{fig:scenario_2_results}
}
\subfloat[Navigation scenario -- Move right.]{\includegraphics[width=0.25\linewidth]{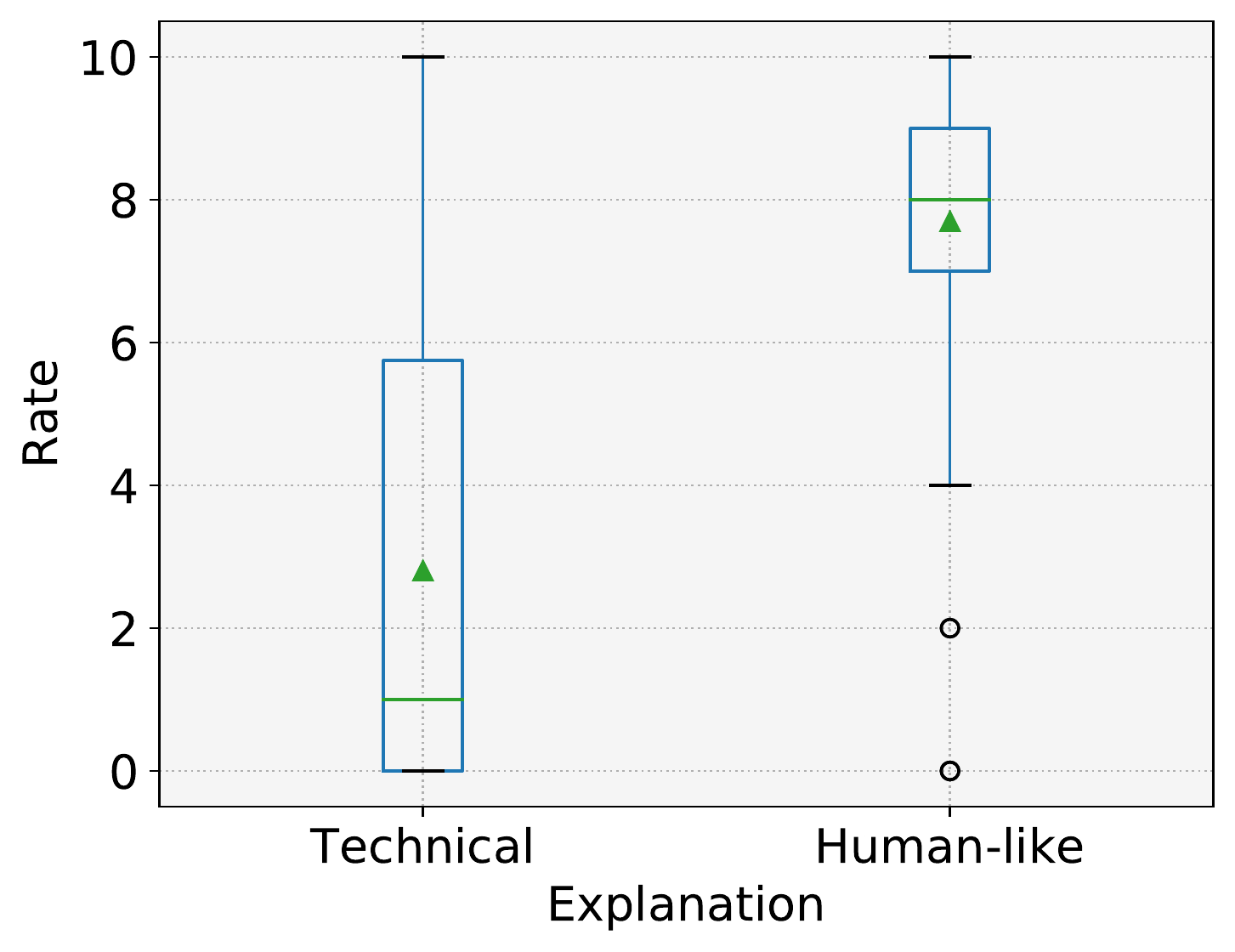} 
\label{fig:scenario_3_results}
} 
\subfloat[Navigation scenario -- Move straight.]{\includegraphics[width=0.25\linewidth]{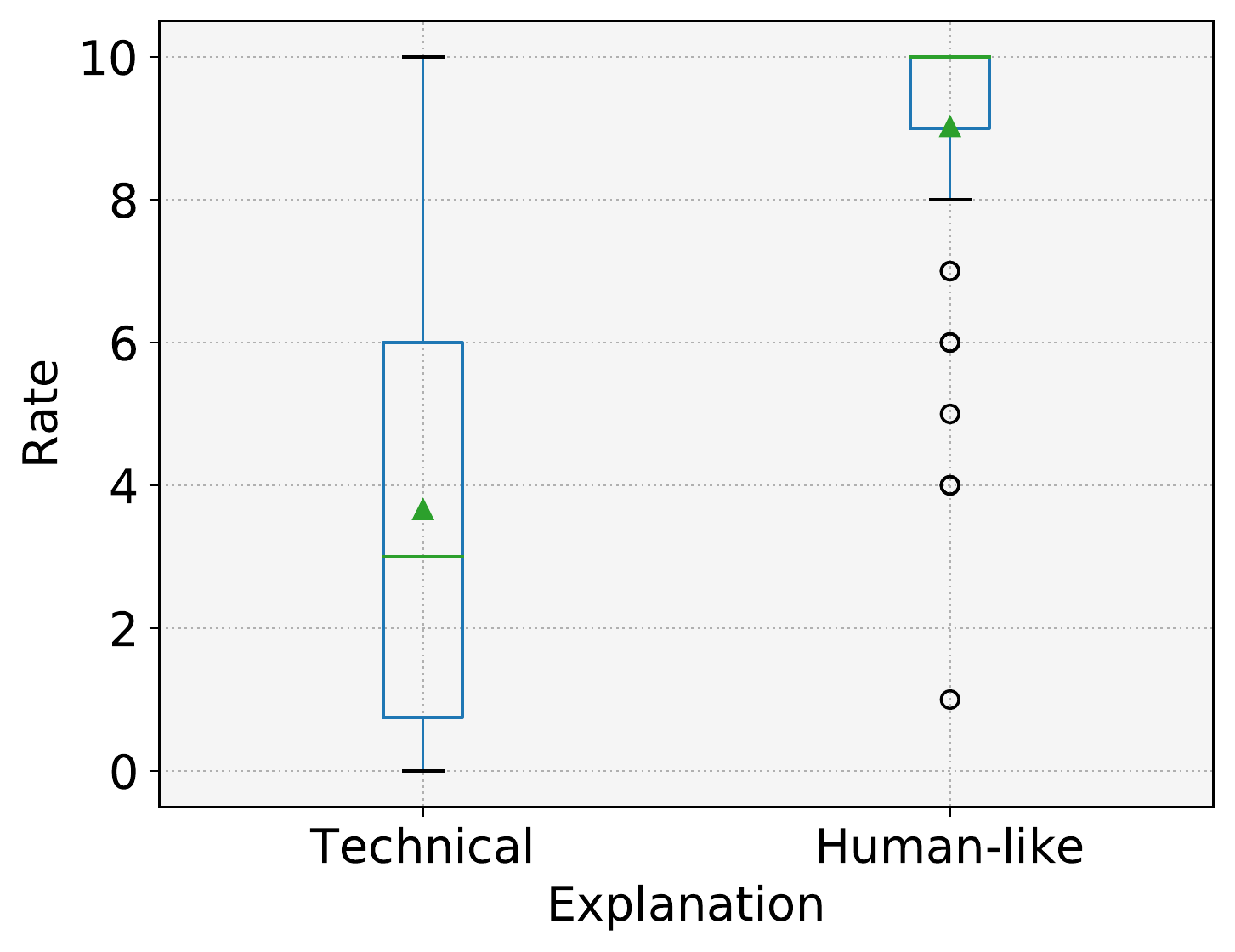} 
\label{fig:scenario_4_results}
}
\\
\subfloat[Robot arm -- Move to the right.]{\includegraphics[width=0.25\linewidth]{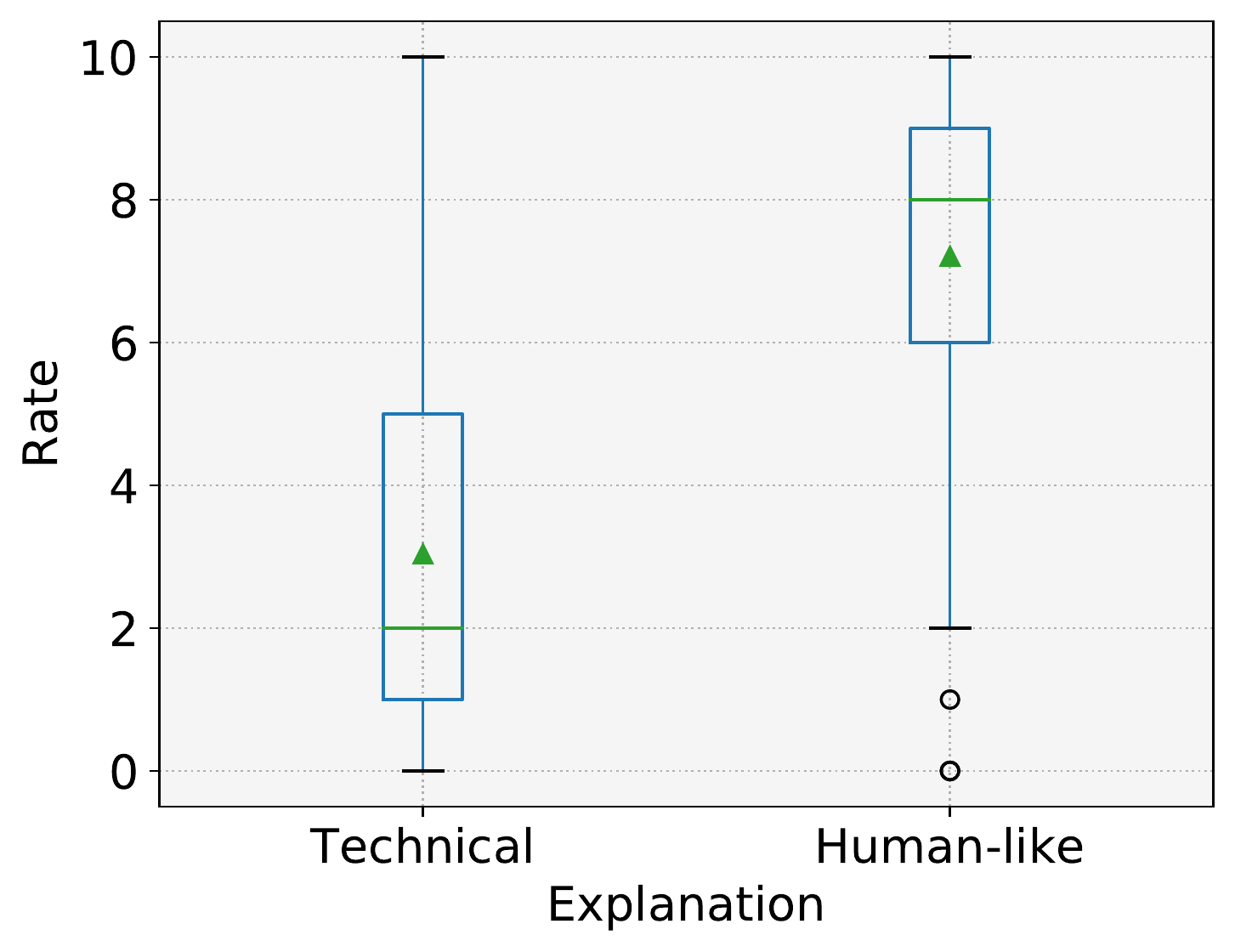} 
\label{fig:scenario_5_results}
}
\subfloat[Robot arm -- Grab an object.]{\includegraphics[width=0.25\linewidth]{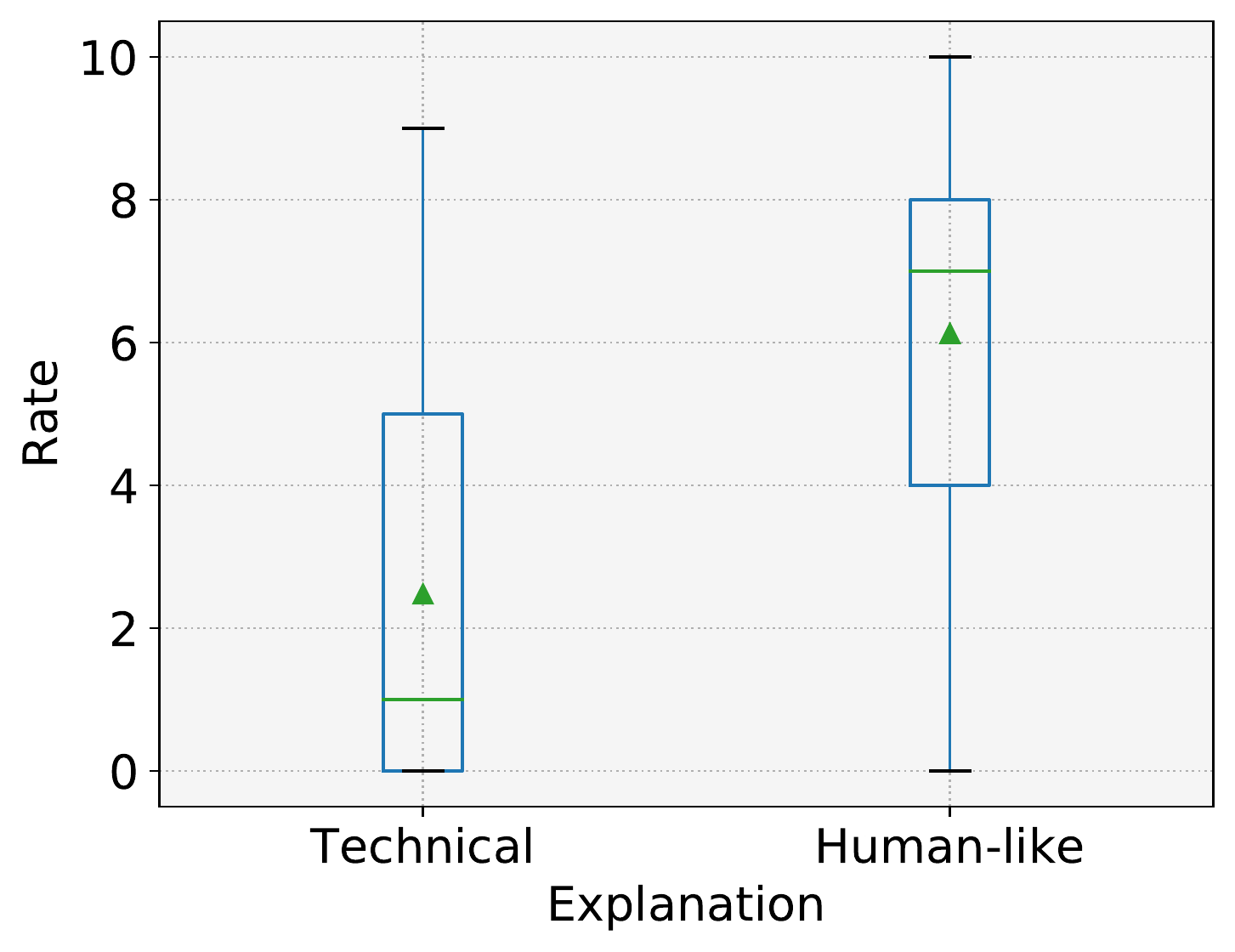} 
\label{fig:scenario_6_results}
} 
%\hspace{1cm}
\subfloat[Real-world Nao -- Move straight.]{\includegraphics[width=0.25\linewidth]{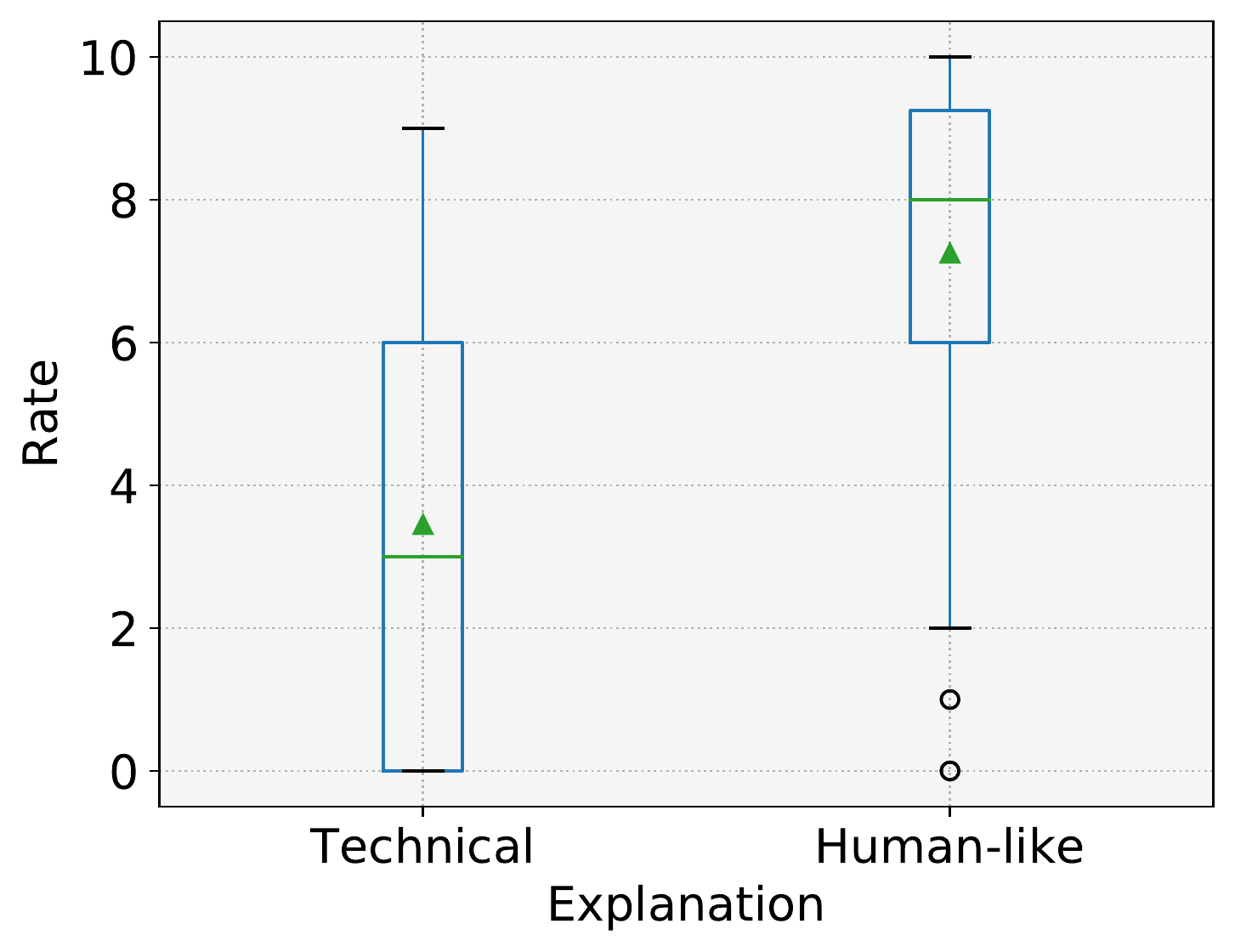} 
\label{fig:scenario_7_results}
}
\subfloat[Real-world Nao -- Move to the left.]{\includegraphics[width=0.25\linewidth]{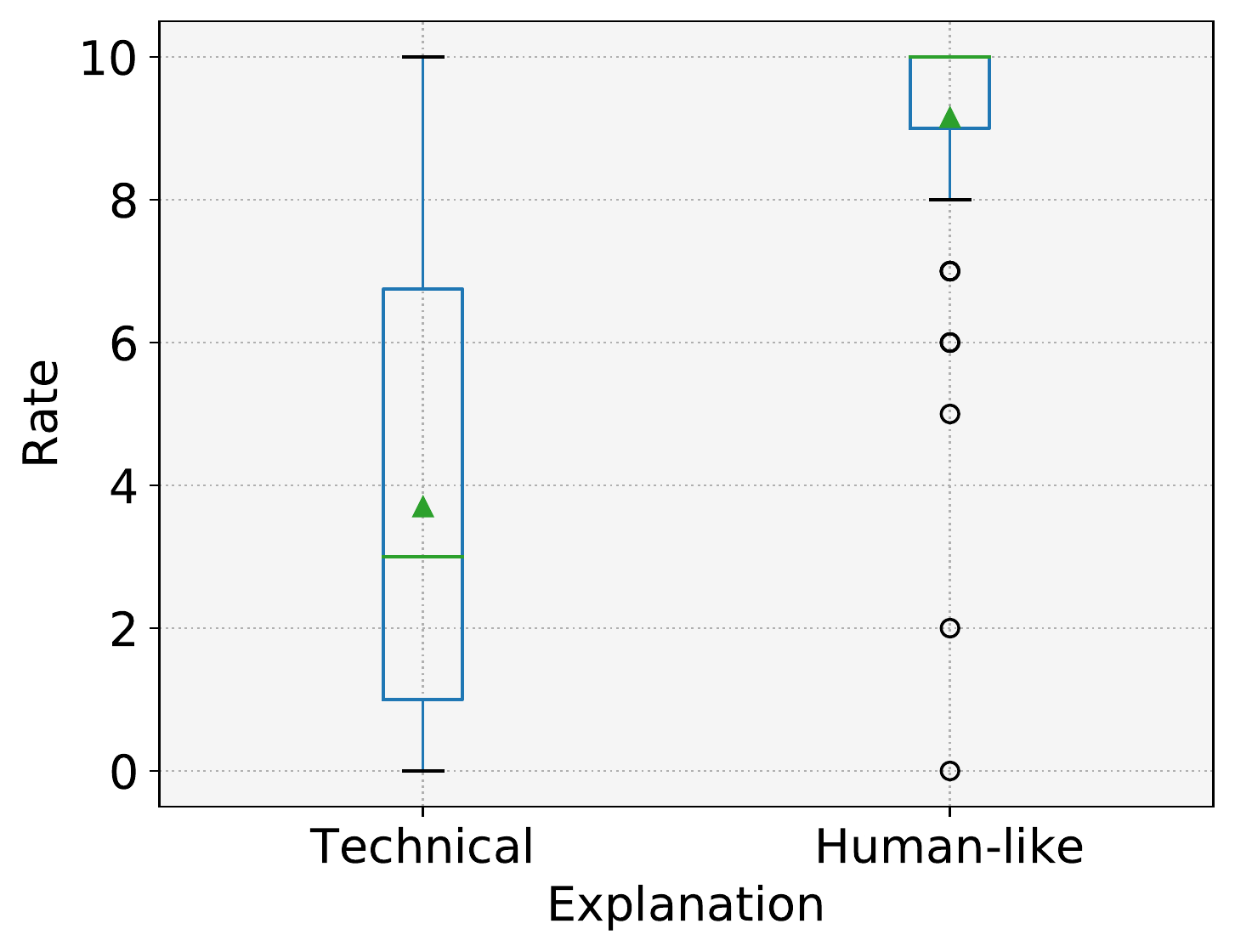} 
\label{fig:scenario_8_results}
} 
\caption{Results obtained for all robotic scenarios. Results show an overall preference for probability-based explanations. The mean evaluation (represented by the green triangle) for the human-like explanations using the probability of success outperforms the technical explanation created from Q-values in all cases. Moreover, evaluations of probability-based explanations show much less variability in comparison to technical explanations.}
\label{fig:results}
\end{figure*}

%As mentioned in Section~\ref{sec:demographic}, demographic questions were asked before starting the survey.
%These questions and their possible answers are shown in Table~\ref{tab:prequestions}.
Demographic questions were asked before starting the survey. % (see Table~\ref{tab:prequestions}). 
%Overall, participants are keen on artificial intelligence within popular culture, e.g., movies, TV series, books, etc. 
%On average, they rated this question as 6.82 ($SD=2.32$).
Most of the participants reported no professional experience in machine learning.
Only 33 participants ($18.03\%$) reported professional experience and 150 participants ($81.97\%$) did not.
Most participants rated their own level of expertise in machine learning as low.
They reported on average a level of expertise of 3.29 ($SD=2.96$).
Figure~\ref{fig:expertise} shows the number of answers for each possible level of expertise reported.
%In terms of previous experience participating in machine learning studies, 57 ($31.15\%$) reported having previously participated, while 126 ($68.85\%$) did not.
%From the ones with previous participation, 32 ($56.14\%$) reported 1 to 2 times, 18 ($31.58\%$) reported 3 to 4 times, and only 7 ($12.28\%$) reported 5 or more times.
Therefore, overall, the participants exhibit very little experience with machine learning methods. 
Additionally, a set of optional questions were asked after the survey.
The gender of the participants was distributed as $59.21\%$ males, $40.21\%$ females, and $0.58\%$ non-binary/third gender.
%A total of 179 participants gave information about their gender, 106 males (59.21\%), 72 females (40.21\%), and 1 non-binary/third gender (0.58\%).
The age of the participants $\in [24, 77]$ was on average 41.10 ($SD=10.95$).
%About the education level of the participants, $16.11\%$ achieved secondary/high school, $65.0\%$ college/university, and $18.89\%$ postgraduate degree.
%The English level reported was $83.33\%$ advanced (C1/C2) or native, $13.89\%$ intermediate (B1/B2), and $2.78\%$ basic (A1/A2).
%Finally, the mean self-perceived level of math knowledge was 5.98 ($SD=2.16$).
%Therefore, given the educational, English, and mathematical level reported by participants, we hypothesize that the provided probability-based explanations should be understood by them.

Figure~\ref{fig:results} shows the results of the survey for each scenario.
In the boxplots, the box represents the distribution between the first and third quartile while the whiskers show the minimum and maximum values.
The green line in the box represents the median, the green triangle is the mean, and the circles are outliers.
It is observed that in all the cases the human-like explanation using the probability of success outperformed the technical explanation built upon the Q-values.
Additionally, it is observed that in all cases the variability between the first and third quartile is much less when using probability-based explanations.

\begin{figure}[htbp]
%\captionsetup[subfigure]{justification=centering}
\subfloat[{Low ML expertise.}]{\includegraphics[width=0.5\linewidth]{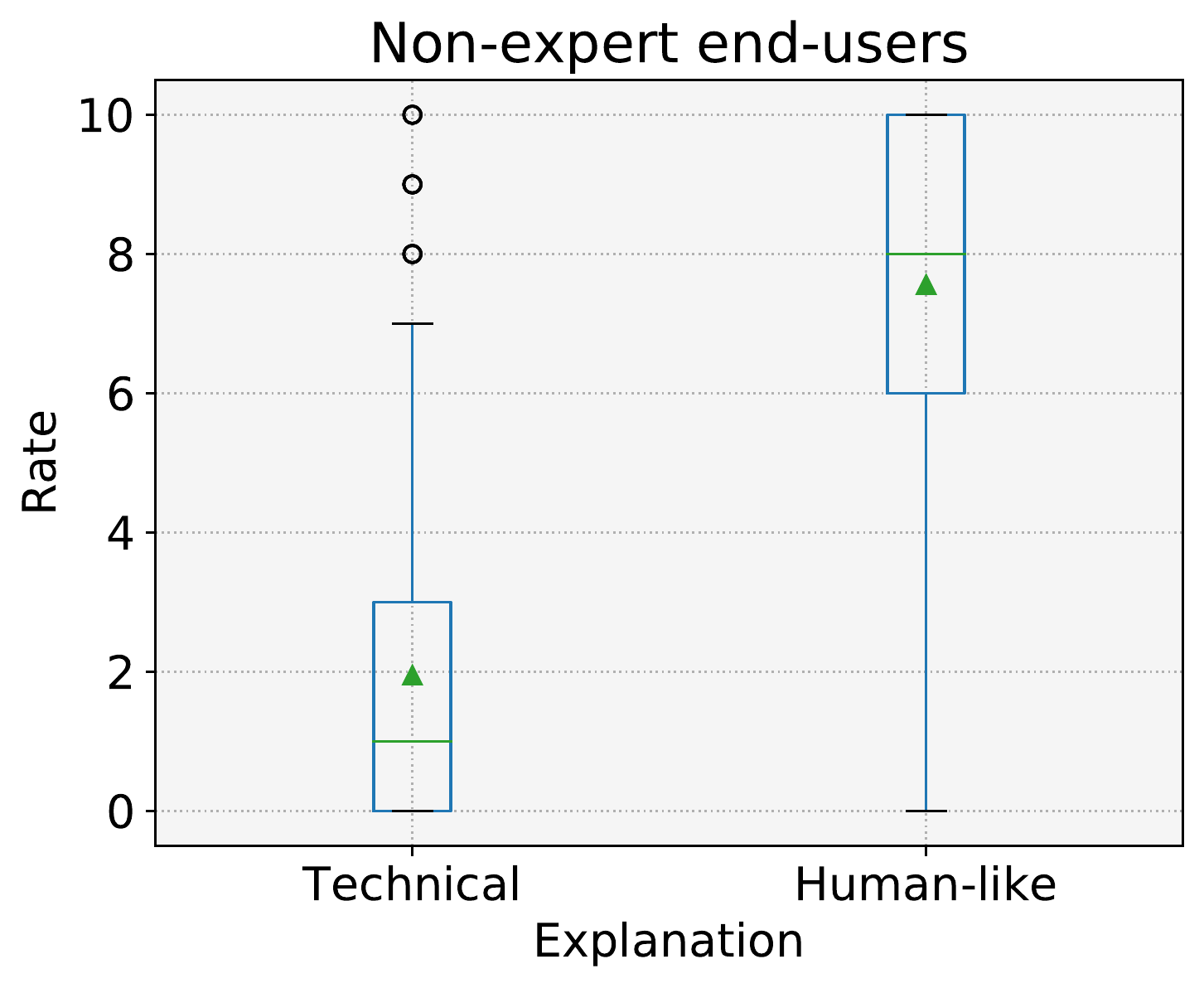} \label{fig:low_scores}} 
%\hspace{1cm}
\subfloat[{High ML expertise.}]{\includegraphics[width=0.5\linewidth]{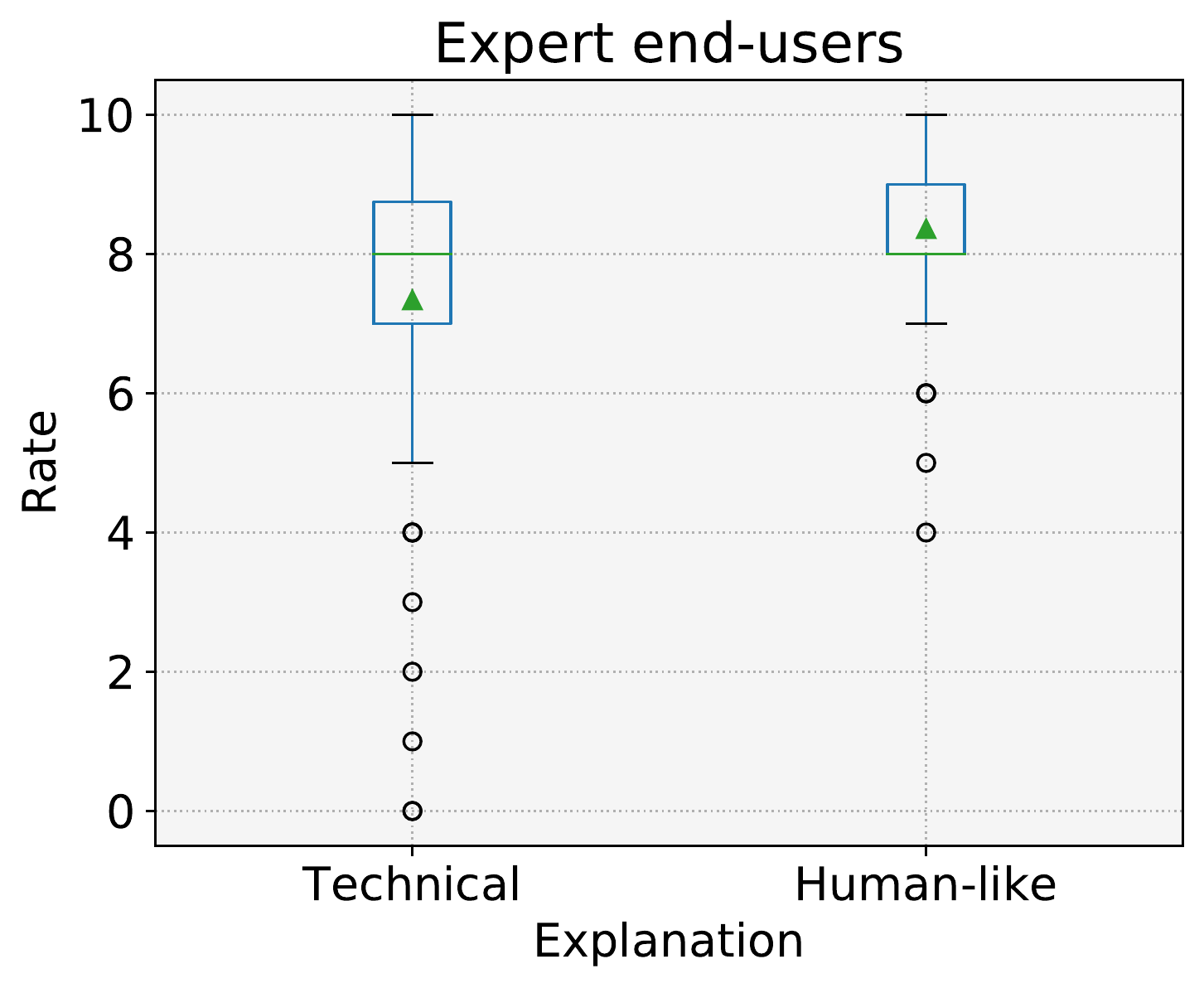} \label{fig:high_score}}
\caption{A comparison of the evaluated explanations between participants with low and high expertise in machine learning. Although both groups prefer probability-based explanations, these are much more meaningful for non-experts.}
\label{fig:low_high_results}
\end{figure}

As the probability-based explanations generated in this work are intended for non-expert end-users, we further analyze the rated explanations in two subgroups, namely, the ones with a low and a high self-reported level of expertise in machine learning.
As mentioned above, Figure~\ref{fig:expertise} shows the detailed self-reported level of expertise.
We built the first subgroup containing the answers from the ones with self-reported low-level of expertise.
We included participants reporting their level as 0, 1, and 2, leading to 84 non-expert end-users.
The second subgroup was built with the participants reporting 8, 9, and 10 as their level of expertise in machine learning, leading to 22 expert end-users.
Figure~\ref{fig:low_high_results} shows a comparison of the evaluation given by expert and non-expert end-users.
On the one hand, Figure~\ref{fig:low_scores} shows that non-expert end-users found probability-based explanations much more useful in comparison to technical explanations.
Considering all scenarios, they rated technical explanations on average as 1.94 ($SD=2.49$) and probability-based explanations on average as 7.56 ($SD=2.80$).
On the other hand, Figure~\ref{fig:high_score} shows that for expert end-users both technical and probability-based explanations are acceptable and understood as they rated both similarly.
They evaluated the technical explanations on average as 7.33 ($SD=2.06$) and probability-based explanations as 8.36 ($SD=1.18$).
Even though in this group both explanations are similarly evaluated, they still rated the probability-based explanations as being slightly better.
This comparison is important since it shows that in particular non-expert end-users indeed favor explanations built with the probability of success, i.e., the probability of success is for them an acceptable basis in order to explain the robot's behavior. 
Therefore, considering that most work has focused on developing explainability for technicians it is important that explanations may also be designed for non-expert end-users specifically as well.

\begin{figure}[htbp]
%\captionsetup[subfigure]{justification=centering}
\subfloat[{Non-experts -- Technical.}]{\includegraphics[width=0.5\linewidth]{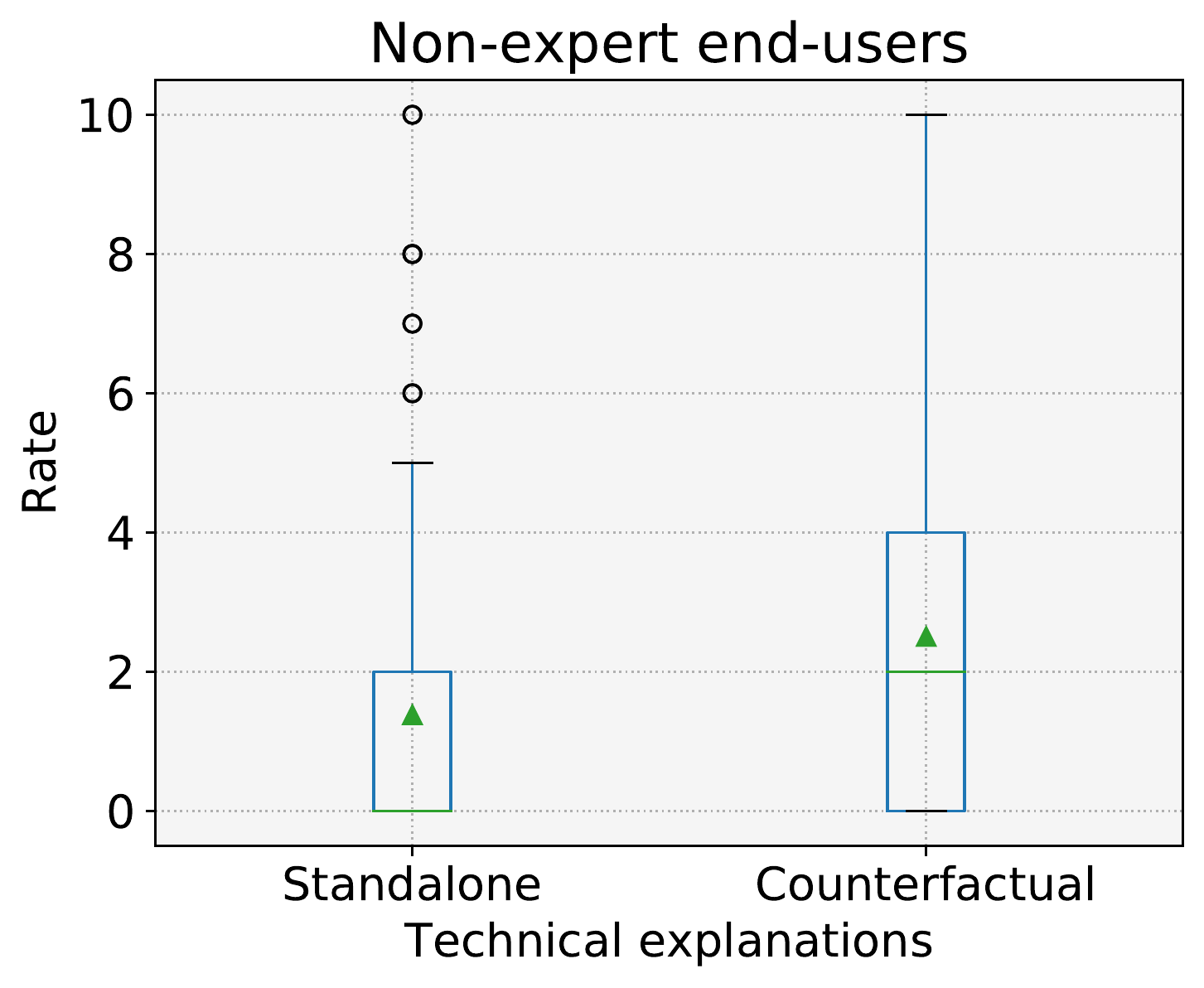} \label{fig:low_scores_technical}} 
%\hspace{1cm}
\subfloat[{Non-experts -- Human-like.}]{\includegraphics[width=0.5\linewidth]{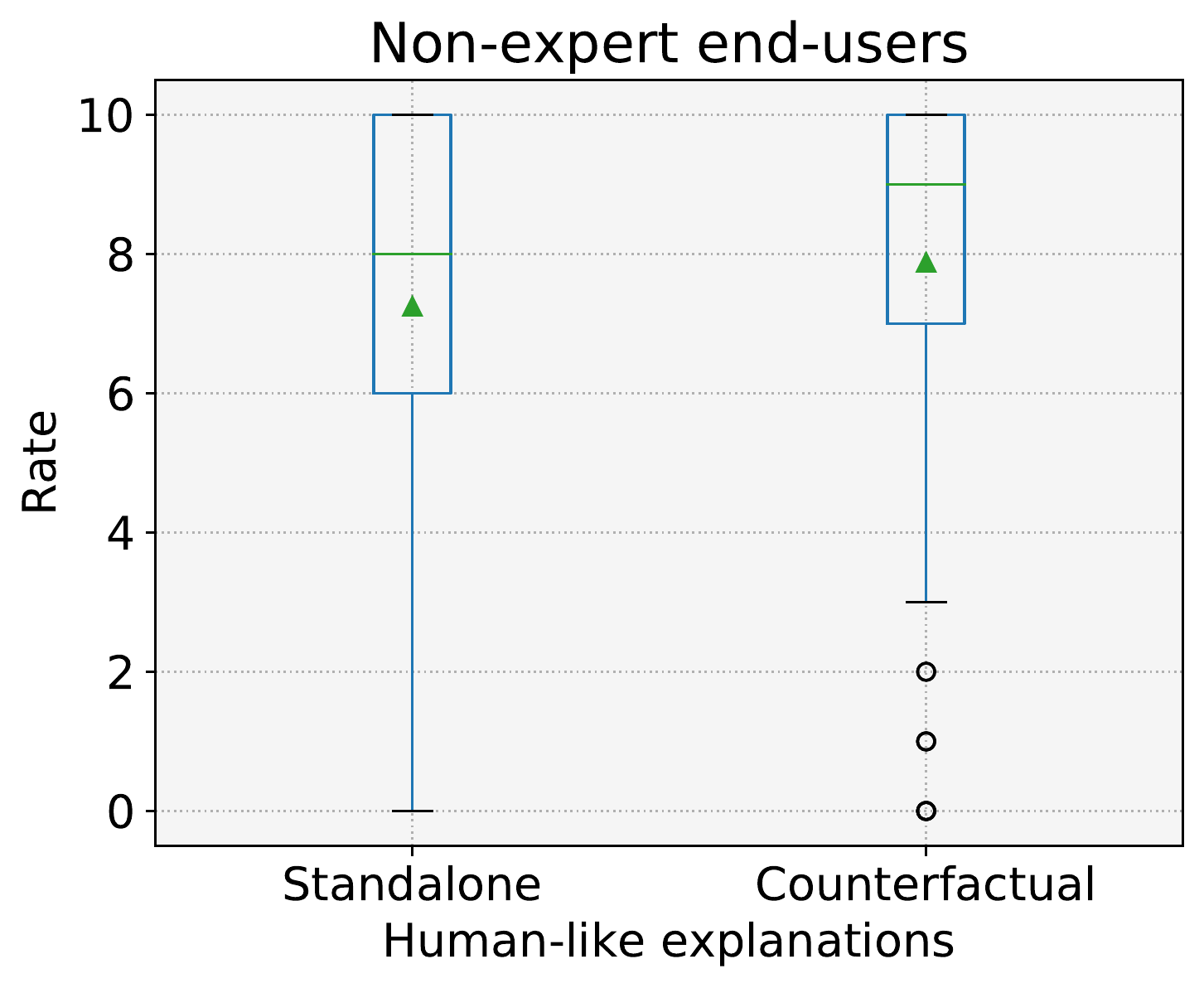} \label{fig:low_score_human}}
\\
\subfloat[{Experts -- Technical.}]{\includegraphics[width=0.5\linewidth]{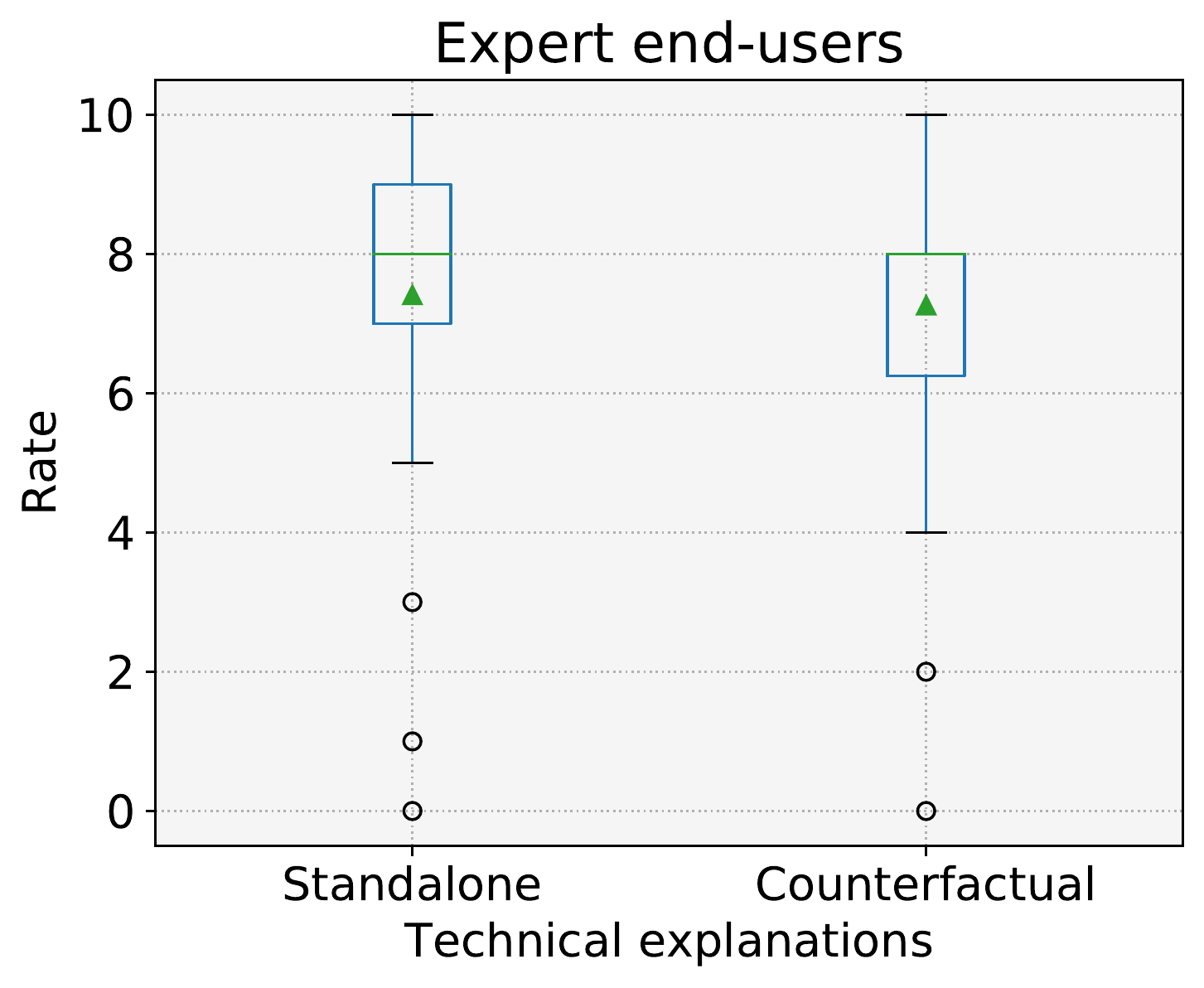} \label{fig:high_scores_technical}} 
%\hspace{1cm}
\subfloat[{Experts -- Human-like.}]{\includegraphics[width=0.5\linewidth]{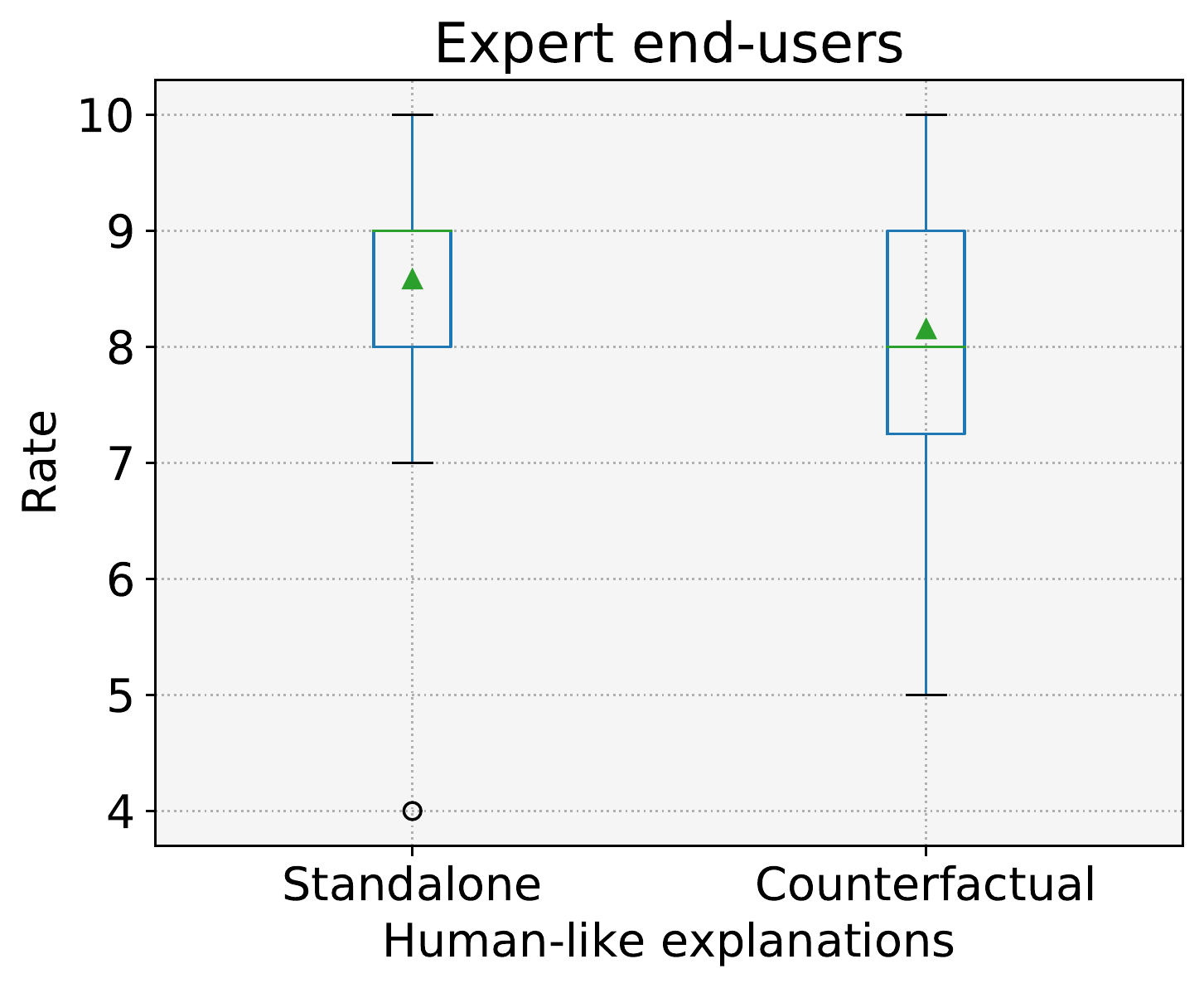} \label{fig:high_score_human}}
\caption{A comparison of standalone and counterfactual explanations. Experts rated better standalone explanations while non-experts preferred counterfactual explanations.}
\label{fig:standalone_counterfactual_results}
\end{figure}

Further analysis on the use of standalone and counterfactual explanations shows a further difference between expert and non-expert users.
As depicted in Figure~\ref{fig:standalone_counterfactual_results}, non-expert end-users tended to favor counterfactual explanations over standalone.
They rated on average standalone technical explanations as 1.38 ($SD=2.15$) and counterfactual technical explanations as 2.50 ($SD=2.68$), (see Figure~\ref{fig:low_scores_technical}).
Rating probability-based explanations, they assigned on average 7.25 ($SD=2.93$) to standalone explanations and 7.88 ($SD=2.63$) to counterfactual explanations (Figure~\ref{fig:low_score_human}).
Differently, expert-end users favored standalone explanations over counterfactuals. 
On average, they rated as 7.41 ($SD=2.06$) standalone technical explanations and as 7.26 ($SD=2.08$) counterfactual technical explanations (Figure~\ref{fig:high_scores_technical}).
Finally, they evaluated the use of probability-based explanations as 8.58 ($SD=1.16$) for standalone and 8.15 ($SD=1.17$) for counterfactual.
Therefore, this is an additional difference in evaluating explanations between expert and non-expert end-users.
While the experts preferred standalone, technical explanations, non-experts preferred counterfactual, probability-based explanations.

\begin{table*}[htbp]
%\begin{table*}[t]
\begin{center}
  \caption{Statistical significance $t$-test for all robotic scenarios. The $p$-values obtained for all scenarios are lower than the hypothesized $\alpha$ level showing a statistically significant difference.}
  \label{tab:test}
  \begin{tabular}{|p{3.0cm}|p{2.2cm}|p{2cm}|p{1cm}|p{2.1cm}|p{1.0cm}|p{2.1cm}|}
  %\hline
  \cline{4-7}

  %\begin{tabular}{ll}\toprule
    \multicolumn{3}{l}{} & \multicolumn{2}{|l|}{\textbf{Technical explanation}} & \multicolumn{2}{|l|}{\textbf{Human-like explanation}} \\ \hline
    \textbf{Scenario} & \textbf{Action} & \textbf{Explanation} & \textbf{$t$-score} & \textbf{p-value} & \textbf{$t$-score} & \textbf{p-value} \\ \hline
    Island scenario & Go east & Standalone & -8.8035 & \num{2.2647E-14} & 9.0122 & \num{8.7327E-15}\\    
    Island scenario & Go south & Counterfactual & -2.8061 & \num{5.9594E-3} & 14.7552 & \num{9.6789E-28} \\ 
    Navigation scenario & Move to the right & Standalone & -7.5695 & \num{1.4515E-11} & 12.7742 & \num{1.8256E-23} \\
    Navigation scenario & Move straight & Counterfactual & -5.1787 & \num{1.0569E-06} & 24.7916 & \num{1.2983E-46} \\
    Robot arm & Move to the right & Standalone & -6.4895 & \num{3.2438E-09} & 9.4656 & \num{4.5575E-16} \\
    Robot arm & Grab an object & Counterfactual & -9.0557 & \num{8.6214E-15} & 4.7980 & \num{4.9822E-06} \\
    Real-world Nao & Move straight & Standalone & -5.0203 & \num{2.0946E-06} & 8.3413 & \num{2.3598E-13} \\
    Real-world Nao & Move to the left & Counterfactual & -4.6840 & \num{8.6200E-06} & 24.4152 & \num{7.0040E-47} \\
    \hline
  \end{tabular}
\end{center}
\end{table*}

Finally, in order to test the statistical significance from the data collected, we have performed a $t$-test for all scenarios.
Our null hypothesis is that the scores are random in a normal distribution. 
Given the score $\in [0, 10]$, we assumed a mean $\mu = 5$ for the scores randomly distributed, and a significance level $\alpha = 0.05$.
The obtained $t$-scores and $p$-values are shown in Table~\ref{tab:test}.
It is observed that in all scenarios for both technical and probability-based explanations all $p$-values are lower than \num{5.9594E-3} and, hence, lower than the hypothesized $\alpha$ level, indicating statistical significant difference.
Therefore, we can reject the null hypothesis and conclude that the scores were not randomly distributed.

\section{Conclusions}
In this paper, we have presented a human study on explainability using technical explanations and human-like explanations built on the probability of success to complete the task.
Participants observed four robotic scenarios in which two actions were taken from an initial situation leading to a change in the robot's state.
After performing each action, the robot offered an explanation of its behavior to be evaluated by end-users.

Results obtained show that participants significantly favor explanations created from the probability of success over technical explanations using Q-values.
Furthermore, human-like explanations are rated with much less variance in comparison to technical explanations.
Additional experiments show that counterfactual probability-based explanations are particularly well evaluated in a subgroup of participants with low expertise in machine learning, referred to as non-expert end-users.
In contrast, a subgroup of expert end-users evaluated both explanations similarly well, even though probability-based standalone explanations were slightly better evaluated than technical or counterfactual ones.
Therefore, we observed that the use of the probability of success is a suitable basis in order to produce human-like explanations understandable for non-expert end-users.

%Interestingly, the subgroup of non-expert end-users evaluated the level of usefulness for technical explanations near 2 (out of 10), although they are likely meaningless for them.
%We hypothesize an `Emperor's New Clothes' situation~\cite{andersen1968emperor}, i.e., participants might try to show they understand the explanation even without technical expertise.
%However, this might also be produced by some deviations from participants not understanding the survey questions or from the survey platform.
As future work, we are interested in performing additional in-person evaluations with both simulated and real-world robotic scenarios.
Additionally, in this study, we have evaluated goal-driven explanations, nevertheless, a complete explanation should consider both goal-driven and state-based aspects.
The explainability system should be able to discriminate what aspect, or to what extent, to use in different situations. 
These mixed explanations must also be rated by non-expert end-users.
Moreover, although trust should be the ultimate goal of an explanation, in this work we are not yet evaluating it. 
As we are planning to extend this work in order to compare state-based and goal-driven explanations, we want to make use of a multi-item validated scale~\cite{hoffman2018metrics}, e.g., goodness criteria or test of comprehension, and evaluate trust by means of questions testing post-hoc comprehension of the explanation.

%%%%%%%%%%%%%%%%%%%%%%%%%%%%%%%%%%%%%%%%%%%%%%%%%%%%%%%%%%%%%%%%%%%%%%%%

\section*{Acknowledgment}
This work was partially founded by Deakin University through the Peer-Review Early Career Researchers Support Scheme (PRESS), the Commonwealth of Australia (represented by the Defence Science and Technology Group) through a Defence Science Partnerships agreement, the Australian Government Research Training Program (RTP) Stipend, RTP Fee-Offset Scholarship through Federation University Australia, and Universidad Central de Chile under the research project CIP2020013.

%\section*{References}
\bibliographystyle{ieeetr}
\balance
\bibliography{biblio}

\end{document}